# Provable Sparse Tensor Decomposition

Will Wei Sun[*] Junwei Lu[†] Han Liu[‡] Guang Cheng[§]

**Abstract**

We propose a novel sparse tensor decomposition method, namely Tensor Truncated Power (TTP) method, that incorporates variable selection into the estimation of decomposition components. The sparsity is achieved via an efficient truncation step embedded in the tensor power iteration. Our method applies to a broad family of high dimensional latent variable models, including high dimensional Gaussian mixture and mixtures of sparse regressions. A thorough theoretical investigation is further conducted. In particular, we show that the final decomposition estimator is guaranteed to achieve a local statistical rate, and further strengthen it to the global statistical rate by introducing a proper initialization procedure. In high dimensional regimes, the obtained statistical rate significantly improves those shown in the existing non-sparse decomposition methods. The empirical advantages of TTP are confirmed in extensive simulated results and two real applications of click-through rate prediction and high-dimensional gene clustering.

## 1 Introduction

Tensor as a multi-dimensional generalization of matrix has been widely used in industry, e.g., Carroll and Chang (1970); Bro and Kiers (2003), and is being actively studied in the community of machine learning (Karatzoglou et al., 2010; Rendle and Schmidt-Thieme, 2010; Zheng et al., 2010; Chi and Kolda, 2012; Liu et al., 2013) and statistics (Zhou et al., 2013; Yang and Dunson, 2013; Yuan and Zhang, 2014). In particular, significant progress has been made toward tensor decomposition (Signoretto et al., 2014; Anandkumar et al., 2014b), which has shown great success in personalized recommendation. Traditional recommendation systems are mainly based on the user-item matrix, whose entry represents each user's behavior on a particular item. To incorporate the contextual information into the analysis, we need to consider a user-item-context tensor. For example, in online advertising, users' click behaviors on a jacket advertisement in the winter could be very different from those in the summer. In this example, the goal of recommendation systems is to recommend to a user the most suitable advertisement in different temporal circumstances such that this user has a large chance to click it. We refer readers to Kolda and Bader (2009a) for a thorough overview of tensor decompositions and their applications.

[*]Research Scientist, Yahoo Research, Sunnyvale, 94087, Email: sunweisurrey@yahoo-inc.com. Part of this research work was conducted when he was a PhD candidate at Purdue.

[†]Department of Operations Research and Financial Engineering, Princeton University, Princeton, NJ 08544, Email: junweil@princeton.edu.

[‡]Assistant Professor, Department of Operations Research and Financial Engineering, Princeton University, Princeton, NJ 08544, Email: hanliu@princeton.edu. Partially supported by NSF CAREER Award DMS-1454377, NSF IIS 1546482-BIGDATA, NIH R01MH102339, NSF IIS1408910, NSF IIS1332109, and NIH R01GM083084.

[§]Associate Professor, Department of Statistics, Purdue University, West Lafayette, IN 47906, Email: chengg@purdue.edu. Partially supported by NSF CAREER Award DMS-1151692, DMS-1418042, Simons Fellowship in Mathematics and a grant from Office of Naval Research. Guang Cheng was on sabbatical at Princeton while part of this work was carried out; he would like to thank the Princeton ORFE department for its hospitality. All authors would like to thank the editors, the associate editor, and three anonymous reviewers for their helpful comments and suggestions which led to a much improved presentation.

As far as we are aware, most existing tensor decomposition results are established in the non-sparse regime where the decomposition components include all features. In many applications, the tensor contains many zeros and the decomposition components are very sparse. For example, in the online advertising example, the probability of a click event is very tiny and the observed user-item-context tensor contains many zeros, i.e., no click. In addition, in high dimensional tensor cases, many features in the components essentially contain no information about the tensor structure, and thus the performance of tensor structure recovery may not be desirable. Moreover, the interpretability of tensor decomposition is inevitably impeded by including non-relevant features. At last, in many applications, it is important to identify which features are crucial. It has been shown that tensor decomposition is applicable for clustering (Anandkumar et al., 2014b). In high-dimensional gene clustering problem, recognizing the relevant genes for clustering is of great interest for scientific discovery. Hence, a more appropriate method that can simultaneously perform tensor decomposition and select informative features is in need.

In this paper, we propose a new sparse tensor decomposition method called Tensor Truncated Power (TTP) that encourages the sparsity of each decomposition component by incorporating a truncation step into the tensor power iteration step. Specifically, in each iteration, the decomposition components are first updated via a tensor power method (Lathauwer et al., 2000; Anandkumar et al., 2014c), and then truncated to only preserve the entries of $s$ largest magnitudes. Here the parameter $s$ is much smaller than the maximal dimension $d$ in all modes, and can be tuned in a data-driven manner. This truncation step efficiently imposes the desirable sparsity of the decomposition components, and hence significantly improves the statistical rate as shown later. Moreover, we provide a provable sparse SVD initialization procedure, which can eventually lead to the global statistical rate of our final estimator.

In theory, we establish both local and global statistical rates of the proposed method. In particular, for an observed noisy tensor with a perturbation error $\mathcal{E}$, we denote its sparse norm as $\eta(\mathcal{E}, d_0)$ with $d_0$ the maximum number of nonzero entries in the decomposition components of the true tensor; see (3.1) for details. Let $s \geq d_0$ be the maximum number of nonzero elements of the estimated decomposition components in our algorithm and let the rank (defined in (2.2)) of the true tensor be $K$. Given an appropriate initialization with an estimation error $\epsilon_0$, our TTP method requires only $O(\log(\epsilon_0/\epsilon_R))$ steps to achieve the desirable statistical rate $\epsilon_R$, where[1]

$$\epsilon_R = O\Big(\underbrace{\eta(\mathcal{E}, d_0 + s)}_{\text{Sample Error}} + \underbrace{\sqrt{K}/d_0}_{\text{Model Error}}\Big).$$

[1]For two sequence $a_n, b_n$, we say $a_n = O(b_n)$ if there exists some positive constant $C_0$ and sufficiently large $n$ such that $a_n \leq C_0 b_n$, on the other hand, we say $a_n = \Omega(b_n)$ if $b_n = O(a_n)$.

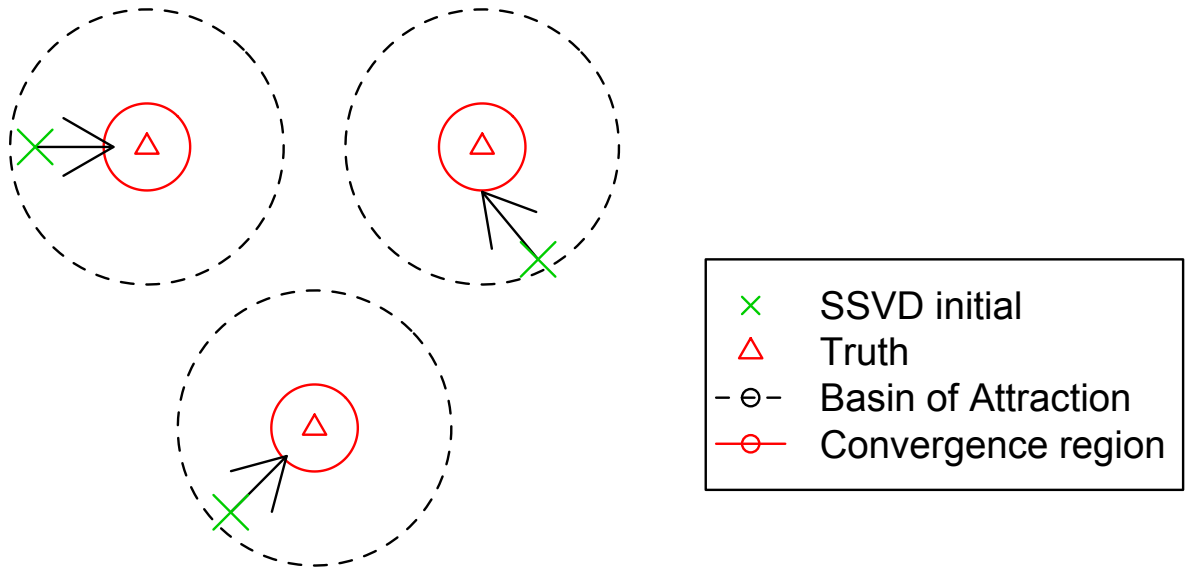

Figure 1: An illustration of our theoretical analysis. The global statistical rate is shown by first proving that the sparse SVD (SSVD) initialization produces an estimator in the basin of attraction, and then carefully quantifying the estimation error at each update step.

This statistical rate can be decomposed into two parts, where the sample error captures the noise level of the observed tensor and the model error measures the complexity in decomposing the true tensor. In high-dimensional regimes, this rate significantly improves those developed in the non-sparse tensor decomposition methods (Anandkumar et al., 2014c), see Section 3.1 for detailed discussions. Figure 1 illustrates our theoretical analysis. When the initialization is not far away from the true parameters, i.e., the so-called "basin of attraction", then our TTP estimator is guaranteed to continuously move toward the convergence region around the true parameters.

A by-product of our TTP method is to solve high-dimensional latent variable models. In order to illustrate the applicability, we employ our method to two statistical problems: high-dimensional clustering (Hsu and Kakade, 2013) and mixture of sparse regressions (Chaganty and Liang, 2013). Finally, extensive experiments are implemented to backup the theoretical developments and demonstrate the superior performance of our procedure.

### 1.1 Comparison with Related Work

A related work on tensor decomposition is the robust tensor decomposition method with non-sparse decomposition components proposed by Anandkumar et al. (2014c). Under certain conditions, they prove that their method is able to recover the decomposition with an error rate $O(\eta(\mathcal{E}, d) + \sqrt{K}/d)$ when the dimensions of all decomposition components equal to $d$. In the high-dimensional regimes where $d_0 \ll d$, the above error is dominated by the sample error $\eta(\mathcal{E}, d)$, which is significantly larger than our sample error $\eta(\mathcal{E}, d_0 + s)$. In order to address high dimensionality, one key ingredient of our method is a new truncation step built upon their robust tensor decomposition method to encourage the sparsity structure of the decomposition components. This additional truncation step demands more challenging technical analysis than those in Anandkumar et al. (2014c). In particular, we need to carefully characterize the impact of the intermediate sparse update on the estimation error in each iteration step. See Section 2.3 for more discussions.

Another line of research focuses on the convex relaxation of the low rank tensor decomposition and completion problem. For example, the convex relaxation is achieved by a generalized trace norm (Romera-Paredes and Pontil, 2013), a tensor Schatten 1-norm (Liu et al., 2014; Gu et al., 2014), or a novel tensor nuclear norm (Yuan and Zhang, 2014). Kolda and Bader (2009b); Plantenga et al. (2015) and Kolda (2015) propose a straightforward optimization method to the symmetric real-valued or nonnegative tensor decomposition. However, all these works mainly focus on low rank tensor recovery and their decomposition components are generally non-sparse.

In sparse tensor decomposition, Morup et al. (2008) develops a sparse non-negative Tucker decomposition approach by matricizing the tensor and employing the $\ell_1$ norm penalty as the sparsity constraint. However, as commented in Allen (2012), in the high-dimensional case, sparse optimization after matrization is computationally intensive and requires a large amount of computer memory. Instead, Allen (2012) directly adds an $\ell_1$ penalty on the decomposition vectors in the rank-1 best approximation optimization problem and solves it via alternative soft thresholding update, while Liu et al. (2012) suggests to solve the sparse non-negative tensor factorization via a coordinate descent method. Although these sparse procedures show good empirical performance in variable selection, no theoretical analysis on these sparse estimators is available. Chi and Kolda (2012) propose a nonnegative decomposition algorithm for tensors whose entries are sparse and their theoretical analysis requires that the tensors are generated from a Poisson distribution. To the best of our knowledge, our TTP method is the first sparse tensor decomposition method with guaranteed local and global statistical rates.

### 1.2 Notation

The following notation is adopted throughout this paper. Denote $[d] = \{1, \ldots, d\}$. For a matrix $\mathbf{A} = (a_{ij}) \in \mathbb{R}^{d \times d}$, we denote $\|\mathbf{A}\|$ as its spectral norm. We define $\mathbf{A}_j$ as the $j$-th column and $\mathbf{A}^j$ as the $j$-th row of $\mathbf{A}$. Furthermore, $\mathbf{A}_{\backslash j} \in \mathbb{R}^{d \times (d-1)}$ is a submatrix of $\mathbf{A}$ with its $j$-th column removed, and $\mathbf{A}^{\backslash j} \in \mathbb{R}^{(d-1) \times d}$ is $\mathbf{A}$ with its $j$-th row removed. We denote the $d \times d$ identity matrix as $\mathbf{I}_d$ or $\mathbf{I}$ when no confusion arises. For a vector $\mathbf{v} = (\mathbf{v}_1, \ldots, \mathbf{v}_d)^\top \in \mathbb{R}^d$, $\|\mathbf{v}\|$ refers to its Euclidean norm and $\|\mathbf{v}\|_0$ denotes the number of nonzero entries of $\mathbf{v}$. For an index set $\mathcal{I} \subseteq [d]$, we define $\mathbf{v}_{\mathcal{I}}$ as the vector whose $i$-th entry is equal to $\mathbf{v}_i$ if $i \in \mathcal{I}$, and zero otherwise. Let $\text{supp}(\mathbf{v})$ be the index set of its nonzero entries. Let $\mathbb{1}(\mathcal{A})$ be the indicator function which equals 1 when the event $\mathcal{A}$ is true and 0 otherwise. We denote $\circ$ to be the outer product between vectors. Throughout this paper, we use $C_0, C_1, \ldots$ to denote generic absolute constants, whose values may vary from line to line.

### 1.3 Paper Organization

The rest of the article is organized as follows. Section 2 introduces our sparse tensor decomposition method TTP and its implementation. The local and global statistical rates of the proposed TTP method are established in Section 3. Section 4 describes practical selections of tuning parameters. Section 5 presents the simulation results, followed by a real application in Section 6. The online supplementary material explains affiliated steps in our main algorithm, discusses applications of the TTP method to high-dimensional latent variable models, and includes additional experimental results and all technical proofs for the theoretical developments.

## 2 Sparse Tensor Decomposition

In this section, we introduce some preliminary background on tensors, and then propose our sparse tensor decomposition method as well as its efficient implementation.

### 2.1 Preliminary

Consider a third-order tensor $\mathcal{T} \in \mathbb{R}^{d_1 \times d_2 \times d_3}$. We adopt the tensor notation in the review article by Kolda and Bader (2009a). In tensors, a *fiber* refers to a higher order analogue of matrix row and column. A fiber is obtained by fixing all but one of the indices of the tensor. For a third-order tensor $\mathcal{T}$, the mode-1 fiber is given by $[\mathcal{T}]_{:,j,l}$, mode-2 fiber by $[\mathcal{T}]_{i,:,l}$ and mode-3 fiber by $[\mathcal{T}]_{i,j,:}$. We similarly define a *slice* of a tensor by fixing all but two of the indices. For instance, the slice along mode-1 is given as $[\mathcal{T}]_{i,:,:}$.

We next define the vector product of a tensor. For a vector $\mathbf{u}^{(k)} \in \mathbb{R}^{d_k}$ with $k = 1, 2, 3$ and a tensor $\mathcal{T} \in \mathbb{R}^{d_1 \times d_2 \times d_3}$, we define the mode-1, mode-2, and mode-3 vector product as,

$$\mathcal{T} \times_1 \mathbf{u}^{(1)} := \sum_{i \in [d_1]} \mathbf{u}_i^{(1)} [\mathcal{T}]_{i,:,:}; \quad \mathcal{T} \times_2 \mathbf{u}^{(2)} := \sum_{j \in [d_2]} \mathbf{u}_j^{(2)} [\mathcal{T}]_{:,j,:}; \quad \mathcal{T} \times_3 \mathbf{u}^{(3)} := \sum_{\ell \in [d_3]} \mathbf{u}_\ell^{(3)} [\mathcal{T}]_{:,:,\ell},$$

which are the multilinear combinations of the tensor slices. We also define the multilinear combination of the tensor mode-1 fibers and the multilinear combination of the tensor entries as

$$\mathcal{T} \times_2 \mathbf{u}^{(2)} \times_3 \mathbf{u}^{(3)} := \sum_{j,\ell} \mathbf{u}_j^{(2)} \mathbf{u}_\ell^{(3)} [\mathcal{T}]_{:,j,\ell}; \quad \mathcal{T} \times_1 \mathbf{u}^{(1)} \times_2 \mathbf{u}^{(2)} \times_3 \mathbf{u}^{(3)} := \sum_{i,j,\ell} \mathbf{u}_i^{(1)} \mathbf{u}_j^{(2)} \mathbf{u}_\ell^{(3)} [\mathcal{T}]_{i,j,\ell}.$$

Similar definitions apply to $\mathcal{T} \times_1 \mathbf{u}^{(1)} \times_2 \mathbf{u}^{(2)}$ and $\mathcal{T} \times_1 \mathbf{u}^{(1)} \times_3 \mathbf{u}^{(3)}$. We define the spectral norm of a tensor $\mathcal{T}$ as $\|\mathcal{T}\| := \sup_{\|\mathbf{u}\|=\|\mathbf{v}\|=\|\mathbf{w}\|=1} |\mathcal{T} \times_1 \mathbf{u} \times_2 \mathbf{v} \times_3 \mathbf{w}|$ and its Frobenius norm as $\|\mathcal{T}\|_F := \sqrt{\sum_{i,j,\ell} [\mathcal{T}]_{i,j,\ell}^2}$.

### 2.2 Sparse Tensor Decomposition Method

Tensor decomposition aims to express a tensor as the sum of rank one tensors. Specifically, a tensor $\mathcal{T} \in \mathbb{R}^{d_1 \times d_2 \times d_3}$ is said to have a rank $K$ if it can be written as the sum of $K$ rank-1 tensors, that is $\mathcal{T} = \sum_{i \in [K]} w_i \mathbf{a}_i \circ \mathbf{b}_i \circ \mathbf{c}_i$, where $w_i \in \mathbb{R}$ and $\mathbf{a}_i \in \mathbb{R}^{d_1}, \mathbf{b}_i \in \mathbb{R}^{d_2}, \mathbf{c}_i \in \mathbb{R}^{d_3}$. Here, we assume $\mathbf{a}_i, \mathbf{b}_i, \mathbf{c}_i$ to be unit vectors, since otherwise the normalized terms can be incorporated in the coefficient $w_i$. Given an observed tensor $\widehat{\mathcal{T}}$, which can be written as $\widehat{\mathcal{T}} = \mathcal{T} + \mathcal{E}$ with $\mathcal{T}$ being the true tensor and $\mathcal{E}$ being the error tensor, we can recover its low rank decomposition by minimizing $\|\widehat{\mathcal{T}} - \sum_{i \in [K]} w_i \mathbf{a}_i \circ \mathbf{b}_i \circ \mathbf{c}_i\|_F$ subject to constraints on $w_i, \mathbf{a}_i, \mathbf{b}_i$, and $\mathbf{c}_i$ (Carroll and Chang, 1970; Bro and Kiers, 2003).

In the simplest case where $K = 1$, the single-factor tensor decomposition solves $\|\widehat{\mathcal{T}} - w \mathbf{a} \circ \mathbf{b} \circ \mathbf{c}\|_F$ subject to $\|\mathbf{a}\| = \|\mathbf{b}\| = \|\mathbf{c}\| = 1$ and $w > 0$, whose solution is given by Allen (2012) as,

$$\widehat{\mathbf{a}} = \text{Norm}(\widehat{\mathcal{T}} \times_2 \mathbf{b} \times_3 \mathbf{c}), \ \widehat{\mathbf{b}} = \text{Norm}(\widehat{\mathcal{T}} \times_1 \mathbf{a} \times_3 \mathbf{c}), \ \widehat{\mathbf{c}} = \text{Norm}(\widehat{\mathcal{T}} \times_1 \mathbf{a} \times_2 \mathbf{b}), \tag{2.1}$$

where $\text{Norm}(\mathbf{v}) = \mathbf{v}/\|\mathbf{v}\|$ is a normalization operator on a vector $\mathbf{v}$. This procedure provides an iterative coordinate update procedure for the single-factor tensor decomposition. To compute all the $K$ decomposition components, one can apply this single-factor procedure sequentially to the residual tensor left after subtracting previously recovered ones. The single-factor tensor decomposition procedure incorporating this deflation strategy is known as the tensor power method (Kolda and Bader, 2009a; Anandkumar et al., 2014c), which is efficient for non-sparse tensor decomposition.

In this paper, we consider a model of sparse and low-rank tensor decomposition. We assume that $\mathcal{T} \in \mathbb{R}^{d_1 \times d_2 \times d_3}$ is sparse and has rank $K$ such that

$$\mathcal{T} = \sum_{i \in [K]} w_i \mathbf{a}_i \circ \mathbf{b}_i \circ \mathbf{c}_i, \quad w_i \in \mathbb{R}, \; \mathbf{a}_i \in \mathbb{S}^{d_1-1}, \mathbf{b}_i \in \mathbb{S}^{d_2-1}, \mathbf{c}_i \in \mathbb{S}^{d_3-1}, \tag{2.2}$$

where $\mathbb{S}^{d-1}(\mathbb{R}) = \{\mathbf{v} \in \mathbb{R}^d \,|\, \|\mathbf{v}\| = 1\}$ and $\|\mathbf{a}_i\|_0 \le d_{01}, \|\mathbf{b}_i\|_0 \le d_{02}, \|\mathbf{c}_i\|_0 \le d_{02}$ for any $i \in [K]$. Moreover, we assume $w_{\max} = w_1 \ge \cdots \ge w_K = w_{\min} > 0$ and assume each $w_i$ to be bounded away from 0 and $\infty$.

In order to recover the sparse and low-rank decomposition components, our sparse tensor decomposition method attaches the following truncation step to each tensor power update. To introduce the truncation step, we first define two relevant operations: for a vector $\mathbf{v} \in \mathbb{R}^d$ and an index set $F \subseteq [d]$, we define Truncate$(\mathbf{v}, F)$ as

$$[\text{Truncate}(\mathbf{v}, F)]_i = \begin{cases} \mathbf{v}_i & \text{if } i \in F \\ 0, & \text{otherwise} \end{cases},$$

and for a scaler $s \le d$, we denote Truncate$(\mathbf{v}, s)$ = Truncate$(\mathbf{v}, \text{supp}(\mathbf{v}, s))$, where supp$(\mathbf{v}, s)$ refers to the set of indices of $\mathbf{v}$ corresponding to its largest $s$ absolute values. Building upon the tensor power update in (2.1), our sparse tensor decomposition method updates the sparse components via

$$\check{\mathbf{a}} = \text{Truncate}(\widehat{\mathbf{a}}, s_1), \;\; \check{\mathbf{b}} = \text{Truncate}(\widehat{\mathbf{b}}, s_2), \;\; \check{\mathbf{c}} = \text{Truncate}(\widehat{\mathbf{c}}, s_3), \tag{2.3}$$

where $s_1 \le d_1, s_2 \le d_2, s_3 \le d_3$ are the corresponding sparsity levels. Then the components are normalized in order to satisfy the unit norm constraint. We refer our method as Tensor Truncated Power (TTP) method. The detailed algorithm is proposed in Section 2.3. As will be shown later, this truncation step achieves desirable variable selection effect and improves the tensor structure recovery performance in the high-dimensional settings.

Note that the above truncation idea has recently shown to be successful in a wide context of high dimensional problems. For example, Yuan and Zhang (2013) employ it to extract the largest sparse eigenvectors of a high dimensional matrix. Wang et al. (2014) apply this strategy to the EM algorithm to produce a sparse solution in high dimensional cases. It is worth noting that our paper is the first one to incorporate this truncation strategy into the tensor decomposition problem.

Our sparse tensor decomposition method is applicable to solve the low rank tensor approximation problem which approximates an observed tensor $\widehat{\mathcal{T}} \in \mathbb{R}^{d_1 \times d_2 \times d_3}$ by a sparse and low rank tensor in (2.2). This is a generalization of penalized matrix decomposition considered in Witten et al. (2009). The sparse tensor approximation aims to minimize $\|\mathcal{T} - \sum_{i \in [K]} w_i \mathbf{a}_i \circ \mathbf{b}_i \circ \mathbf{c}_i\|_F$ subject to the constraints that $w_i \in \mathbb{R}^+, \; \mathbf{a}_i \in \mathbb{S}^{d_1-1}, \mathbf{b}_i \in \mathbb{S}^{d_2-1}, \mathbf{c}_i \in \mathbb{S}^{d_3-1}$ and $\|\mathbf{a}_i\|_0 \le d_{01}, \|\mathbf{b}_i\|_0 \le d_{02}, \|\mathbf{c}_i\|_0 \le d_{02}$. However, solving the optimization problem directly is computationally challenging. We can apply our TTP method to find an approximation to the solution of this optimization problem.

### 2.3 Algorithm

In Algorithm 1 below, we present more implementation details of the proposed TTP method.

---

**Algorithm 1** TTP Method for Sparse Tensor Decomposition

---

**Input:** tensor $\widehat{\mathcal{T}} \in \mathbb{R}^{d_1 \times d_2 \times d_3}$, number of initializations $L$, number of iterations $N$, cardinality vector $(s_1, s_2, s_3)$, rank $K$.

**For** $\tau = 1$ **to** $L$ **Do**

  **Initialize** unit vectors $\widehat{\mathbf{a}}_\tau^{(0)} \in \mathbb{R}^{d_1}, \widehat{\mathbf{b}}_\tau^{(0)} \in \mathbb{R}^{d_2}, \widehat{\mathbf{c}}_\tau^{(0)} \in \mathbb{R}^{d_3}$.

  **For** $t = 1$ **to** $N$: Alternatively update the components $\widehat{\mathbf{a}}_\tau^{(t)}, \widehat{\mathbf{b}}_\tau^{(t)}, \widehat{\mathbf{c}}_\tau^{(t)}$ as

$$\bar{\mathbf{a}}_\tau^{(t)} = \text{Norm}\Big(\widehat{\mathcal{T}} \times_2 \widehat{\mathbf{b}}_\tau^{(t-1)} \times_3 \widehat{\mathbf{c}}_\tau^{(t-1)}\Big);\ \check{\mathbf{a}}_\tau^{(t)} = \text{Truncate}\big(\bar{\mathbf{a}}_\tau^{(t)}, s_1\big);\ \widehat{\mathbf{a}}_\tau^{(t)} = \text{Norm}\big(\check{\mathbf{a}}_\tau^{(t)}\big), \tag{2.4}$$

$$\bar{\mathbf{b}}_\tau^{(t)} = \text{Norm}\Big(\widehat{\mathcal{T}} \times_1 \widehat{\mathbf{a}}_\tau^{(t-1)} \times_3 \widehat{\mathbf{c}}_\tau^{(t-1)}\Big);\ \check{\mathbf{b}}_\tau^{(t)} = \text{Truncate}\big(\bar{\mathbf{b}}_\tau^{(t)}, s_2\big);\ \widehat{\mathbf{b}}_\tau^{(t)} = \text{Norm}\big(\check{\mathbf{b}}_\tau^{(t)}\big), \tag{2.5}$$

$$\bar{\mathbf{c}}_\tau^{(t)} = \text{Norm}\Big(\widehat{\mathcal{T}} \times_1 \widehat{\mathbf{a}}_\tau^{(t-1)} \times_2 \widehat{\mathbf{b}}_\tau^{(t-1)}\Big);\ \check{\mathbf{c}}_\tau^{(t)} = \text{Truncate}\big(\bar{\mathbf{c}}_\tau^{(t)}, s_3\big);\ \widehat{\mathbf{c}}_\tau^{(t)} = \text{Norm}\big(\check{\mathbf{c}}_\tau^{(t)}\big). \tag{2.6}$$

  **End For**

**End For**

**Output:** the cluster centers $(\widehat{\mathbf{a}}_j, \widehat{\mathbf{b}}_j, \widehat{\mathbf{c}}_j), j \in [K]$ by clustering $\Big\{(\widehat{\mathbf{a}}_\tau^{(N)}, \widehat{\mathbf{b}}_\tau^{(N)}, \widehat{\mathbf{c}}_\tau^{(N)}), \tau \in [L]\Big\}$ into $K$ clusters and their corresponding $\widehat{w}_j = \widehat{\mathcal{T}} \times_1 \widehat{\mathbf{a}}_j \times_2 \widehat{\mathbf{b}}_j \times_3 \widehat{\mathbf{c}}_j$.

---

The key step of our TTP procedure is the truncated power updates in (2.4)-(2.6). In each iteration $t$ of the inner loop, when updating the first decomposition component in (2.4), we first compute the non-sparse component $\bar{\mathbf{a}}_\tau^{(t)}$ via classical tensor power method in (2.1), then the truncation step keeps only the top $s_1$ entries in $\bar{\mathbf{a}}_\tau^{(t)}$ and sets the rest entries as zero. Finally, the truncated component is normalized to $\widehat{\mathbf{a}}_\tau^{(t)}$ which ends a full update cycle of the first decomposition component. Our TTP method updates all three components alternatively until a pre-specified maximal number of iterations is achieved. This termination condition can be modified to the case when the changes of components are below some thresholding value, see Section 4.1 for more details.

For each initialization $\tau$, the procedure runs $N$ iterations of updates in (2.4)-(2.6) to generate the converged decomposition components. These procedures are repeated for $L$ different initializations, where the number $L$ is a polynomial function of $K$. A theoretical lower bound of $L$ will be given in Section 3 and a practical choice of $L$ will be provided in Section 4. In the algorithm, we suggest two initialization procedures, one is a sparse SVD initialization and another is a random initialization. The former has a nice theoretical guarantee, while the latter is simple and fast in practice. The detailed initialization procedures are discussed in Section S.1.1 in the supplementary. In the last step, the algorithm clusters the $L$ tuples of $(\widehat{\mathbf{a}}_\tau^{(N)}, \widehat{\mathbf{b}}_\tau^{(N)}, \widehat{\mathbf{c}}_\tau^{(N)})$ into $K$ clusters to output all components. This clustering step will be discussed in details in Section S.1.2 in the supplementary. In practice, the parameters $s_1, s_2, s_3$, and $K$ can be determined via a data-driven tuning procedure and will be discussed in details in Section 4.

It's worth mentioning that our Algorithm 1 is built based on the non-sparse tensor decomposition algorithm proposed by Anandkumar et al. (2014c). In order to address the high dimensionality, one key ingredient of our method is a new truncation step in (2.4)-(2.6) to encourage the sparsity structure of the decomposition components. Despite its simplicity, this additional truncation step demands more challenging technical analysis than those in Anandkumar et al. (2014c). In particular, we need to carefully characterize the impact of the intermediate sparse update on the estimation error in each iteration step. As we will show in the next Section, such truncation step leads to a much improved statistical rate in high-dimensional tensor decomposition scenario. Moreover, different from the sparse decomposition procedure considered in Allen (2012) which imposes $\ell_1$ penalty on each component, our TTP method encourages the sparsity via a direct $\ell_0$ cardinality constraint. Although the $\ell_1$ penalized decomposition procedure in Allen (2012) shows good empirical performance in variable selection, no theoretical analysis of its performance was available. To the best of our knowledge, our TTP method is the first sparse tensor decomposition method with guaranteed both local and global statistical rates.

# 3 Theoretical Analysis

This section establishes both local and global analysis of our TTP method. Specifically, to show the local statistical rate, we require an incoherence condition on the true tensor, a constraint on the perturbation error, as well as an appropriate initialization; to show the global statistical rate, we then employ the local analysis coupled with the error analysis of the newly introduced sparse SVD initialization procedure.

Recall that we focus on the sparse and low-rank tensor decomposition of the true tensor $\mathcal{T} \in \mathbb{R}^{d_1 \times d_2 \times d_3}$ as defined in (2.2). In practice, we usually observe a perturbed tensor $\widehat{\mathcal{T}}$ based on limited samples. That is, given the perturbation error $\mathcal{E}$, the observed tensor $\widehat{\mathcal{T}}$ can be written as $\widehat{\mathcal{T}} = \mathcal{T} + \mathcal{E}$. To quantify the noise level of the error, we define the sparse spectral norm of $\mathcal{E}$ as

$$\eta(\mathcal{E}, d_{01}, d_{02}, d_{03}) := \sup_{\substack{\|\mathbf{u}\|=\|\mathbf{v}\|=\|\mathbf{w}\|=1 \\ \|\mathbf{u}\|_0 \le d_{01}, \|\mathbf{v}\|_0 \le d_{02}, \|\mathbf{w}\|_0 \le d_{03}}} \Big| \mathcal{E} \times_1 \mathbf{u} \times_2 \mathbf{v} \times_3 \mathbf{w} \Big|. \tag{3.1}$$

Here $\eta(\mathcal{E}, d_{01}, d_{02}, d_{03})$ quantifies the perturbation error in a sparse scenario, and in the sparse tensor decomposition case with $d_{01} \ll d_1, d_{02} \ll d_2, d_{03} \ll d_3$, it is much smaller than the spectral norm $\|\mathcal{E}\|$, which equals $\eta(\mathcal{E}, d_1, d_2, d_3)$. Denote $d_0 = \max\{d_{01}, d_{02}, d_{03}\}$, we have $\eta(\mathcal{E}, d_{01}, d_{02}, d_{03}) \le \eta(\mathcal{E}, d_0, d_0, d_0)$ and for simplicity we denote $\eta(\mathcal{E}, d_0) := \eta(\mathcal{E}, d_0, d_0, d_0)$.

Our theoretical analysis studies the sufficient conditions under which the estimated components $(\widehat{\mathbf{a}}_j, \widehat{\mathbf{b}}_j, \widehat{\mathbf{c}}_j)$ from our TTP method converge to the truth $(\mathbf{a}_j, \mathbf{b}_j, \mathbf{c}_j)$ for any $j \in [K]$. Moreover, our analysis quantifies its specific statistical rate. In order to compute the distance between the estimator and the truth, we define the distance measure between two unit vectors $\mathbf{u}, \mathbf{v} \in \mathbb{R}^d$ as

$$D(\mathbf{u}, \mathbf{v}) := \sqrt{1 - (\mathbf{u}^\top \mathbf{v})^2}. \tag{3.2}$$

For unit vectors $\mathbf{u}, \mathbf{v}$, we have $D(\mathbf{u}, \mathbf{v}) \le \min\{\|\mathbf{u} - \mathbf{v}\|, \|\mathbf{u} + \mathbf{v}\|\} \le \sqrt{2} D(\mathbf{u}, \mathbf{v})$. The distance function $D(\mathbf{u}, \mathbf{v})$ resolves the sign issue in the decomposition components since changing the signs of any two components vectors while fixing the third component vector will not affect the generated tensor.

Before presenting the main theorems, we introduce assumptions on the identifiability and incoherence on the true tensor $\mathcal{T}$, which play an important role in our subsequent theoretical analysis.

**Assumption 3.1 (Identifiability).** The tensor decomposition of $\mathcal{T}$ in (2.2) is unique in the sense that if the tensor has another decomposition $\mathcal{T} = \sum_{i\in[K']} w_i'\mathbf{a}_i' \circ \mathbf{b}_i' \circ \mathbf{c}_i'$ with $\mathbf{a}_i' \in \mathbb{S}^{d_1-1}, \mathbf{b}_i' \in \mathbb{S}^{d_2-1}, \mathbf{c}_i' \in \mathbb{S}^{d_3-1}$ and $w_i' \neq 0$, we have $K = K'$ and there must exist a permutation $\sigma$ of $\{1, \ldots, K\}$ such that $w'_{\sigma(i)} = w_i, \mathbf{a}'_{\sigma(i)} = \mathbf{a}_i$, $\mathbf{b}'_{\sigma(i)} = \mathbf{b}_i$ and $\mathbf{c}'_{\sigma(i)} = \mathbf{c}_i$.

**Assumption 3.2 (Incoherence).** The decomposition components are incoherent such that

$$\zeta := \max_{i\neq j}\{|\langle \mathbf{a}_i, \mathbf{a}_j\rangle|, |\langle \mathbf{b}_i, \mathbf{b}_j\rangle|, |\langle \mathbf{c}_i, \mathbf{c}_j\rangle|\} \leq \frac{C_0}{\sqrt{d_0}}, \tag{3.3}$$

with $d_0 = \max\{d_{01}, d_{02}, d_{03}\}$ and for any $j$, $\| \sum_{i\neq j} w_i\langle \mathbf{a}_i, \mathbf{a}_j\rangle\langle \mathbf{b}_i, \mathbf{b}_j\rangle \mathbf{c}_i\| \leq C_1 w_{\max}\sqrt{K}\zeta$. Moreover, matrices $\mathbf{A} := [\mathbf{a}_1, \cdots, \mathbf{a}_K]$, $\mathbf{B} := [\mathbf{b}_1, \cdots, \mathbf{b}_K]$, and $\mathbf{C} := [\mathbf{c}_1, \cdots, \mathbf{c}_K]$ satisfy $\max\{\|\mathbf{A}\|, \|\mathbf{B}\|, \|\mathbf{C}\|\} \leq 1 + C_2\sqrt{K/d_0}$ for some positive constants $C_0, C_1, C_2$.

**Remark 3.3.** Kruskal (1976, 1977) provide the classical condition of the identifiability of tensor decomposition, that is, it is sufficient for the uniqueness of the decomposition in (2.2) if $k_{\mathbf{A}}+k_{\mathbf{B}}+k_{\mathbf{C}} \geq 2K+2$, where $k_{\mathbf{A}}, k_{\mathbf{B}}, k_{\mathbf{C}}$ are the Kruskal ranks [2] of the matrices $\mathbf{A}, \mathbf{B}, \mathbf{C}$. Such a condition requires that the rank is of the same order as the dimension of the tensor. Under the overcomplete case that $K > \max\{d_1, d_2, d_3\}$, Chiantini and Ottaviani (2012) prove that the set of tensors not having a unique tensor decomposition has Lebesgue measure zero and show that the generic identifiability condition holds if $K \leq (d_1 + 1)(d_2 + 1)/16$. Therefore, Assumption 3.1 is satisfied for most of the tensor decomposition problems.

**Remark 3.4.** The incoherence condition can be viewed as a relaxation of the orthogonality of decomposition components. It is originally introduced by Donoho and Huo (2001) and has been widely studied in high-dimensional scenarios, for example, compressed sensing (Candes and Romberg, 2007) and matrix decomposition (Chandrasekaran et al., 2012). In the experiments, we will illustrate that the incoherence condition of Assumption 3.2 holds if the components $\mathbf{a}_i, \mathbf{b}_i, \mathbf{c}_i$ are randomly generated from the unit and sparse space.

## 3.1 Local Statistical Rate

This section introduces our first main theoretical result of local analysis under the assumptions on true tensor, the perturbation error, as well as the initialization.

We start the local analysis by defining the error term. Recall that $\mathcal{E}$ is the tensor of perturbation error, $d_0 = \max\{d_{01}, d_{02}, d_{03}\}$ is the maximal number of nonzero elements in the true decomposition components, $s = \max\{s_1, s_2, s_3\}$ is the maximal number of nonzero elements in the estimated decomposition components from Algorithm 1, and $K$ is the tensor rank. Denote the error

$$\epsilon_R := \frac{2\sqrt{5}}{w_{\min}}\eta(\mathcal{E}, d_0 + s) + \frac{2\sqrt{5}C_1 w_{\max}}{w_{\min}}\sqrt{K}\zeta^2. \tag{3.4}$$

[2]The Kruskal rank of a matrix is the maximal number $k$ such that every $k$ columns of the matrix are linearly independent.

The first term in (3.4) represents the sample error caused by the perturbation tensor $\mathcal{E}$ and the second term is the model error characterized by the incoherent parameter $\zeta$. If the eigenvectors are orthogonal, the incoherent parameter $\zeta = 0$ and the model error in (3.4) disappears. Under Assumption 3.2, the error term has upper bound $2\sqrt{5}(w_{\min})^{-1}[\eta(\mathcal{E}, d_0 + s) + C_0^2 C_1 w_{\max}\sqrt{K}/d_0]$. We also note that when the sparsity $d_0$ is small, Assumption 3.2 is not sufficient for the estimation consistency. However, as long as we have $\sqrt{K}\zeta^2 = o(1)$, the error term is controllable.

The local statistical rate relies on the following initialization condition.

**Assumption 3.5** (**Initialization**)**.** Define the initialization error $\epsilon_0 := \max\{D(\widehat{\mathbf{a}}^{(0)}, \mathbf{a}_j), D(\widehat{\mathbf{b}}^{(0)}, \mathbf{b}_j)\}$ for some $j \in [K]$. We assume that

$$\epsilon_0 \le \gamma := \min\left\{\frac{w_{\min}}{6w_{\max}} - \frac{C_1\sqrt{K}}{d_0}, \frac{w_{\min}}{4\sqrt{5}C_3 w_{\max}} - \frac{2C_0}{C_3\sqrt{d_0}}\left(1 + C_2\sqrt{\frac{K}{d_0}}\right)^2\right\}.$$

Given $\widehat{\mathbf{a}}^{(0)}, \widehat{\mathbf{b}}^{(0)}$, the sparse vector $\widehat{\mathbf{c}}^{(0)}$ is calculated based on (2.6).

We are now ready to introduce our first main theorem on local analysis.

**Theorem 3.6.** (**Local statistical rate**) Consider the model in (2.2) satisfying Assumptions 3.1 and 3.2, and assume $\|\mathcal{T}\| \le C_3 w_{\max}$ and $K = o(d_0^{3/2})$ with $d_0 = \max\{d_{01}, d_{02}, d_{03}\}$. Let $\widehat{\mathcal{T}}$ be an input to Algorithm 1. Assume the perturbation error satisfies $\eta(\mathcal{E}, d_0 + s) \le w_{\min}/6$ and the initialization $(\widehat{\mathbf{a}}^{(0)}, \widehat{\mathbf{b}}^{(0)}, \widehat{\mathbf{c}}^{(0)})$ satisfies Assumption 3.5 for some $j \in [K]$. The solution from the inner loop of Algorithm 1 with $s_i \ge d_{0i}$ for $i = 1, 2, 3$, after $N = \Omega\big(\log(\epsilon_0/\epsilon_R)\big)$ iterations, satisfies with high probability,

$$\max\left\{D(\widehat{\mathbf{a}}^{(N)}, \mathbf{a}_j), D(\widehat{\mathbf{b}}^{(N)}, \mathbf{b}_j), D(\widehat{\mathbf{c}}^{(N)}, \mathbf{c}_j)\right\} \le O(\epsilon_R). \tag{3.5}$$

Moreover, let $\widehat{w} = \widehat{\mathcal{T}} \times_1 \widehat{\mathbf{a}}^{(N)} \times_2 \widehat{\mathbf{b}}^{(N)} \times_3 \widehat{\mathbf{c}}^{(N)}$, then we have $|\widehat{w} - w_j| \le O(\epsilon_R)$ with high probability.

The proof of Theorem 3.6 is shown in Section S.4.1. The key idea is to show that the estimator from our TTP procedure has an error contraction effect in each iteration, which leads to the desirable contraction result after sufficiently many iterations.

Theorem 3.6 establishes the local statistical rate. According to (3.4), the error term $\epsilon_R$ is of the order $O(\eta(\mathcal{E}, d_0 + s) + \sqrt{K}/d_0)$. Here the recovery error $\epsilon_R$ consists of two parts: the first part is the sample error due to the perturbation error of the observed tensor which is unavoidable, and the second part is the model error characterized by the incoherence Assumption 3.2 (that allows the non-orthogonality of decomposition components). Consider the case that there is no perturbation error, i.e., $\mathcal{E} = 0$ and the error $\epsilon_R$ is in the order of $\sqrt{K}/d_0$. In this case, for any fixed $K$, when $d_0$ is large, the incoherent parameter $\zeta$ in Assumption 3.2 becomes small and hence each decomposition component is nearly orthogonal, which leads to a simple tensor decomposition problem; on the other hand, when $d_0$ is small, the structure of decomposition components could be more complex, and hence the recovery error is very large. Moreover, the requirement on the number of iteration $N = \Omega\big(\log(\epsilon_0/\epsilon_R)\big)$ indicates that, if the initialization is appropriate, i.e., $\epsilon_0$ is small, we only need a few steps to achieve a desirable error bound.

**Remark 3.7.** It is worth noting that in the high dimensional regimes, our error rate $\epsilon_R$ significantly improves the rate shown in Anandkumar et al. (2014c). Under certain conditions, Anandkumar et al. (2014c) prove that their method is able to recover the decomposition with an error rate $O(\eta(\mathcal{E}, d) + \sqrt{K}/d)$ when $d_1 = d_2 = d_3 = d$. In the high-dimensional regimes where $d$ is large, this error is dominated by the sample error $\eta(\mathcal{E}, d)$, which is significantly larger than our sample error $\eta(\mathcal{E}, d_0 + s)$. This improvement is further backed up by our experimental studies in Section 5.

If we replace the truncation step Truncate$(\cdot, s)$ in (2.4) - (2.6) of our Algorithm 1 by a soft-thresholding operator $S(\cdot, \rho) = \text{sign}(\cdot)\max(|\cdot| - \rho, \cdot)$ suggested by Allen (2012), we can achieve the following theoretical result similar to the one shown in Theorem 3.6. Therefore, our proof techniques are also applicable to establish the theoretical performance of the lasso penalized sparse tensor decomposition method in Allen (2012).

**Corollary 3.8.** Suppose the same conditions in Theorem 3.6 are satisfied with $s = d_0$. If the tuning parameter in the soft-thresholding operator satisfies $\rho \geq \epsilon_R$ and the minimal absolute value of the nonzero entries of $\{\mathbf{a}_i, \mathbf{b}_i, \mathbf{c}_i\}$ for $1 \leq i \leq K$ is larger than $2\rho$, we have the same rate as (3.5) with high probability.

## 3.2 Global Statistical Rate

In order to show the global statistical rate of our TTP method, we need to quantify the error bound of the SVD-based initialization in Algorithm 3. In particular, based on the theoretical analysis of the initialization, i.e., Lemma S.4.2 in the supplementary, we establish the following global result.

**Theorem 3.9.** (**Global statistical rate**) Consider model in (2.2) satisfying Assumptions 3.1 and 3.2, and assume $\|\mathcal{T}\| \leq C_3 w_{\max}$ and $K = O(d_0)$ with $d_0 = \max\{d_{01}, d_{02}, d_{03}\}$. Let $\widehat{\mathcal{T}}$ be an input to Algorithm 1. Assume the perturbation error satisfies $\eta(\mathcal{E}, d_0 + s) \leq \min\Big\{w_{\min}/6, (w_{\min}/C_5)\sqrt{\log K/s}\Big\}$ for some constant $C_5 > 0$. Let the number of initializations $L = K^{\Omega(\gamma^{-4})}$ with $\gamma$ defined in Assumption 3.5 and the number of iterations $N = \Omega\big(\log(\gamma/\epsilon_R)\big)$. For any $j \in [K]$, the output of our algorithm with $s_i \geq d_{0i}$ for $i = 1, 2, 3$ satisfies

$$\begin{aligned} \max\Big\{D(\widehat{\mathbf{a}}_j, \mathbf{a}_j), D(\widehat{\mathbf{b}}_j, \mathbf{b}_j), D(\widehat{\mathbf{c}}_j, \mathbf{c}_j)\Big\} &\leq O(\epsilon_R), \\ |\widehat{w}_j - w_j| &\leq O(\epsilon_R), \end{aligned}$$

with high probability.

Theorem 3.9 establishes the global statistical rate of the whole procedure by using the sparse SVD as an initialization. Compared to the assumptions in local analysis, Theorem 3.9 requires stronger conditions on $K$ and $\eta(\mathcal{E}, d_0 + s)$. It's worth noting that in general $\gamma$ is a constant and hence the number of initialization $L$ is a polynomial function of $K$. In Section 4, we will discuss how to choose these parameters in practice.

# 4 Practical Choice of Tuning Parameters

In Section 3, for simplicity we assume the number of initializations $L$, the number of iterations $N$, the rank $K$, and the cardinality parameters $s_1, s_2, s_3$ in Algorithm 1 are known. In this section, we discuss how to choose these tuning parameters in practice.

## 4.1 The Number of Initializations and The Number of Iterations

Theorem 3.9 provides a theoretical condition on the number of iterations $L = K^{\Omega(\gamma^{-4})}$, which is a polynomial function of $K$. Based on our extensive experiments in Section 5, we find that in practice it is sufficient to choose $L = \max\{10, K^3\}$.

Moreover, in practice we do not need to specify the number of iterations $N$ in advance, instead we set a termination condition of the truncated power update (2.4)-(2.6) in Algorithm 1 as

$$\max\{\|\widehat{\mathbf{a}}_\tau^{(t)} - \widehat{\mathbf{a}}_\tau^{(t-1)}\|, \|\widehat{\mathbf{b}}_\tau^{(t)} - \widehat{\mathbf{b}}_\tau^{(t-1)}\|, \|\widehat{\mathbf{c}}_\tau^{(t)} - \widehat{\mathbf{c}}_\tau^{(t-1)}\|\} \leq 10^{-4},$$

for each iteration $\tau$. Similar strategy has been used in the non-sparse tensor decomposition algorithm in Anandkumar et al. (2014c). The convergence of the truncated power method has been shown in the sparse matrix decomposition problem by Yuan and Zhang (2013), which can be easily extended to our tensor decomposition case.

## 4.2 Rank and Cardinality Estimation

Our TTP method relies on two key components: the rank $K$ and the cardinality parameters $s_1, s_2, s_3$. It has been shown that exact tensor rank calculation is an NP hard problem (Kolda and Bader, 2009a). In this subsection, following the tuning method in Allen (2012), we provide a BIC-type criterion to estimate these parameters in practice.

Given a pre-specified set of rank values $\mathcal{K}$ and a pre-specified set of cardinality values $\mathcal{S}_1, \mathcal{S}_2, \mathcal{S}_3$, we choose the combination of parameters $(\widehat{K}, \widehat{s}_1, \widehat{s}_2, \widehat{s}_3)$ which minimizes

$$\text{BIC} := \log\left(\frac{\|\mathcal{T} - \sum_{i\in[K]} w_i \mathbf{a}_i \circ \mathbf{b}_i \circ \mathbf{c}_i\|_F^2}{d_1 d_2 d_3}\right) + \frac{\log(d_1 d_2 d_3)}{d_1 d_2 d_3}\left[\sum_{i\in[K]} \left(\|\mathbf{a}_i\|_0 + \|\mathbf{b}_i\|_0 + \|\mathbf{c}_i\|_0\right)\right]. \quad (4.1)$$

The detailed tuning procedure is shown in Algorithm 2. The efficacy of this tuning procedure is evaluated in various experimental results in Section 5.

**Algorithm 2** Tuning Procedure for TTP Method

1: **Input:** tensor $\widehat{\mathcal{T}}$, set of rank values $\mathcal{K}$, set of cardinality values $\mathcal{S}_1, \mathcal{S}_2, \mathcal{S}_3$.
2: **For** each $K \in \mathcal{K}$, $s_1 \in \mathcal{S}_1$, $s_2 \in \mathcal{S}_2$, $s_3 \in \mathcal{S}_3$ **Do**
3: Step 1: Run Algorithm 1 with parameters $K, s_1, s_2, s_3$, and $L = \max\{10, K^3\}$.
4: Step 2: Compute BIC in (4.1) using $\widehat{\mathcal{T}}$ and the decomposed components from Step 1.
5: **End For**
6: **Output:** $(\widehat{K}, \widehat{s}_1, \widehat{s}_2, \widehat{s}_3)$ which minimizes the corresponding BIC.

# 5 Simulation Study

In this section, we demonstrate the mechanism of the proposed TTP method and illustrate its superior performance in sparse tensor recovery.

In order to measure the recovery quality of the tensor decomposition component as well as its weight, we calculate the mean vector estimation error and weight estimation error:

$$\text{mean error} := \frac{1}{3K}\sum_{k=1}^{K}\left\{\|\widehat{\mathbf{a}}_k - \mathbf{a}_k\| + \|\widehat{\mathbf{b}}_k - \mathbf{b}_k\| + \|\widehat{\mathbf{c}}_k - \mathbf{c}_k\|\right\};\ \text{weight error} := \frac{\|\widehat{w} - w\|}{\|w\|}. \quad (5.1)$$

To evaluate the variable selection quality, we compute the true positive rate $\text{TPR} := (\text{TPR}_a + \text{TPR}_b + \text{TPR}_c)/3$ and the false positive rate $\text{FPR} := (\text{FPR}_a + \text{FPR}_b + \text{FPR}_c)/3$, where

$$\text{TPR}_a := \frac{1}{K}\sum_{k=1}^{K}\frac{\sum_i \mathbb{1}([\mathbf{a}_k]_i \neq 0, [\widehat{\mathbf{a}}_k]_i \neq 0)}{\sum_i \mathbb{1}([\mathbf{a}_k]_i \neq 0)};\ \text{FPR}_a := \frac{1}{K}\sum_{k=1}^{K}\frac{\sum_i \mathbb{1}([\mathbf{a}_k]_i = 0, [\widehat{\mathbf{a}}_k]_i \neq 0)}{\sum_i \mathbb{1}([\mathbf{a}_k]_i = 0)},$$

and $\text{TPR}_b, \text{TPR}_c, \text{FPR}_b, \text{FPR}_c$ are defined analogously.

## 5.1 Illustration of TTP Method

Denote a sparse and unit space $\mathbb{S}^{d-1}(d_0) := \{\mathbf{x} \in \mathbb{R}^d : \|\mathbf{x}\| = 1, \|\mathbf{x}\|_0 \le d_0\}$. In order to generate the random decomposition vectors $\mathbf{a}_k, \mathbf{b}_k, \mathbf{c}_k$ from $\mathbb{S}^{d_1-1}(d_{10}), \mathbb{S}^{d_2-1}(d_{20}), \mathbb{S}^{d_3-1}(d_{30})$, respectively for each $k \in [K]$, we first generate i.i.d. standard Gaussian entries of three components $\mathbf{A} := [\mathbf{a}_1, \cdots, \mathbf{a}_K]$, $\mathbf{B} := [\mathbf{b}_1, \cdots, \mathbf{b}_K]$, and $\mathbf{C} := [\mathbf{c}_1, \cdots, \mathbf{c}_K]$, then truncate each column of $\mathbf{A}, \mathbf{B}, \mathbf{C}$ with cardinality parameters $d_{01}, d_{02}, d_{03}$ correspondingly, and finally normalize each column and aggregate the coefficients as $w_k$.

We first empirically check that the tensor based on randomly generated components satisfies the incoherence Assumption 3.2. We set the rank $K = 10$, the dimension $d_1 = d_2 = d_3 = d$, and the true sparsity level $d_{01} = d_{02} = d_{03} = d_0$, and fix $d_0 = d/10$ for each dimension $d \in [10, 1000]$. In this sparse tensor scenario, we compute the incoherence condition $\zeta$ defined in Assumption 3.2 with respect to $d_0$. Figure 2 shows that the incoherence $\zeta$ decays polynomially in $d_0$, which demonstrates the polynomial bound in Assumption 3.2.

We next evaluate the efficacy of the tuning procedure in Algorithm 2. We consider the example with dimensions $d_1 = 40, d_2 = 30, d_3 = 20$, true cardinality parameters $d_{01} = 8, d_{02} = 6, d_{03} = 4$, and the rank $K = 3$. In Algorithm 2, we let the pre-specified set of rank values be $\mathcal{K} = \{1, \ldots, 8\}$ and the pre-specified set of cardinality values be $\mathcal{S}_j = d_j \times 10^{\{-2,-1.9,\ldots,0\}}$ for $j = 1, 2, 3$. Among all the combinations, our tuning algorithm outputs $\widehat{K} = 3$ and $s_j = d_j \times 10^{0.2}$. In the left plot of Figure 3, we show the plot of BIC as a function of $K$ given the selected cardinality parameter, and in the middle and right plots of Figure 3, we compute the estimation error in recovering tensor decomposition component and evaluate the TPR/FPR with respect to Ratio $= 10^{\{-2,-1.9,\ldots,0\}}$ given the selected true $K$. Clearly, the selected ratio from our tuning algorithm leads to the minimal estimation error and the best variable selection performance.

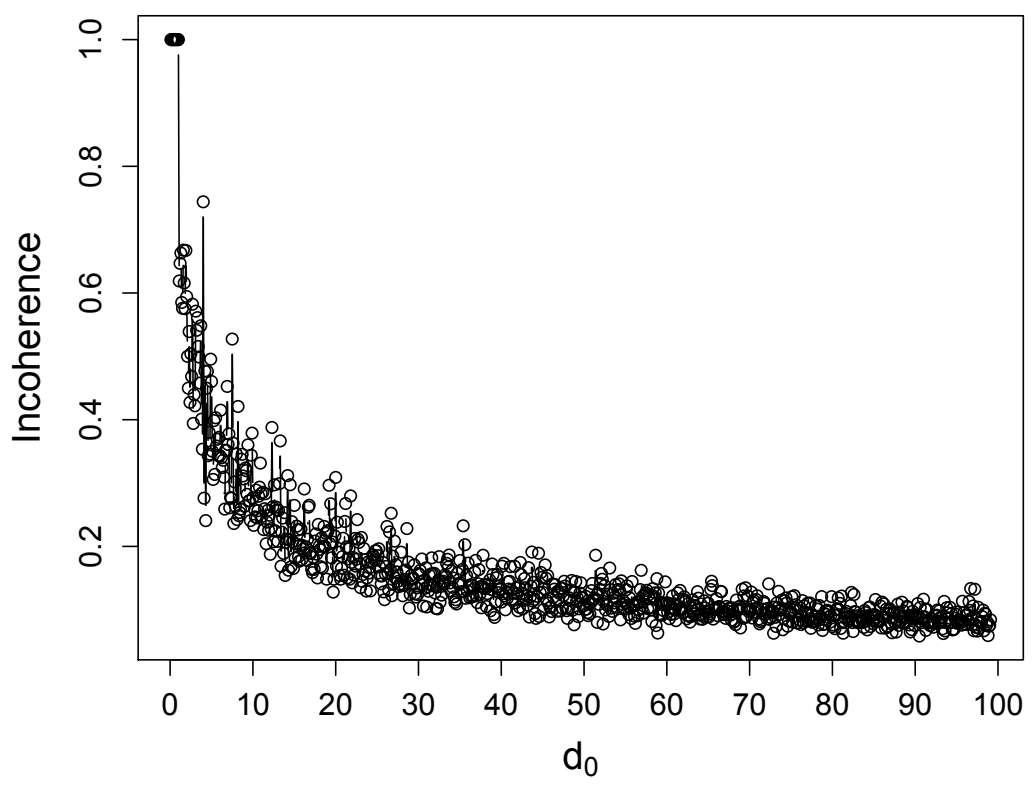

Figure 2: Incoherence $\zeta$ of randomly generated sparse components versus sparse cardinality $d_0$.

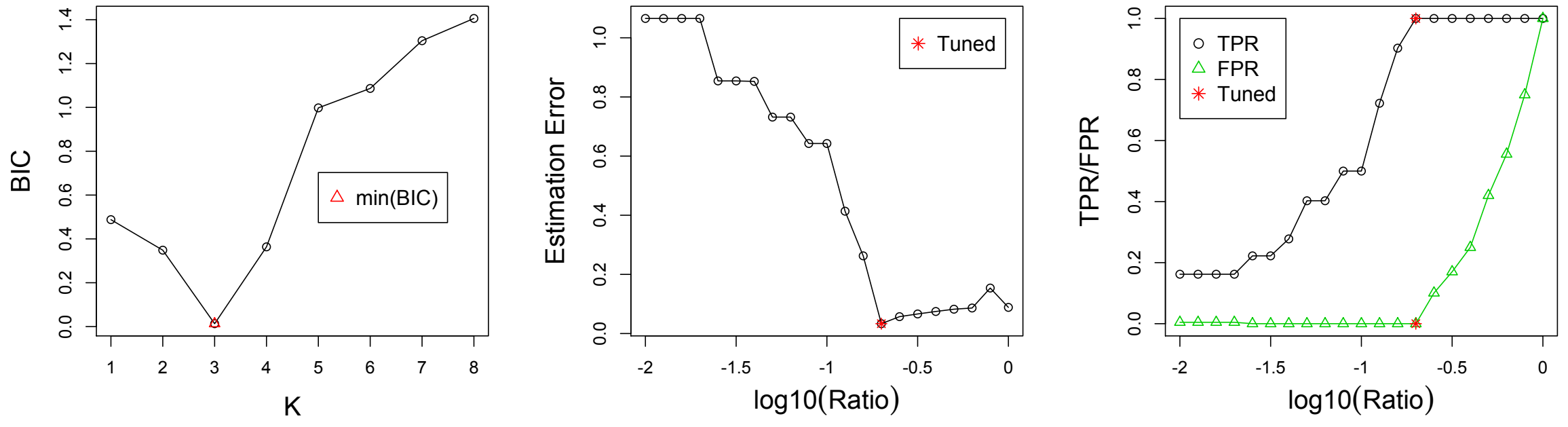

Figure 3: Illustration of the tuning procedure in Algorithm 2. The left plot shows BIC as a function of $K$, the middle plot and the right plot show the estimation error and variable selection performance with respect to Ratio= $10^{\{-2,-1.9,\ldots,0\}}$ given the selected true $K$.

## 5.2 Comparison with Competitive Methods

In this subsection, we compare our TTP method with two competitors: the non-sparse tensor decomposition method in Anandkumar et al. (2014c) and the lasso penalized sparse tensor decomposition method in Allen (2012). In our TTP method, the random initialization in Section S.1.1 is employed and the tuning parameters are selected via Algorithm 2 with $\mathcal{S}_j = d_j \times 10^{\{-2,-1.9,\ldots,0\}}$ $(j = 1, 2, 3)$. For a fair comparison, the tuning parameters $\lambda_1, \lambda_2, \lambda_3$ in the lasso penalized method are also tuned via BIC with the same pre-specified set of parameters $10^{\{-2,-1.9,\ldots,0\}}$. In all three methods, we use the same true rank $K$ such that the comparison is not mixed with the choice of $K$.

In all simulations, we fix the true cardinality $d_{0j} = 0.2d_j$ $(j = 1, 2, 3)$ and vary values of the dimension and the rank. In particular, we consider four scenarios: **I:** $d_1 = 1000, d_2 = 10, d_3 = 10, K = 1$; **II:** $d_1 = 1000, d_2 = 10, d_3 = 10, K = 2$; **III:** $d_1 = 1000, d_2 = 100, d_3 = 10, K = 1$; **IV:** $d_1 = 1000, d_2 = 100, d_3 = 10, K = 2$. Note that in scenarios **III** and **IV**, three dimensions $d_1, d_2, d_3$ are different and three cardinality parameters $d_{01}, d_{02}, d_{03}$ are also different.

We calculate the averaged mean estimation errors, averaged weight estimation errors, and averaged variable selection criterions TRP/FPR over 30 random replications. As shown in Table 1, in all scenarios our TTP method achieves significantly better performance than the non-sparse tensor decomposition method in Anandkumar et al. (2014c) in both estimation accuracy and variable selection performance. In particular, our improvement in sparse mean vector estimation is more than 50% over all scenarios, which indicates the importance of variable selection in sparse tensor decomposition. In addition, compared to the lasso penalized method in Allen (2012), our method obtains better performance in recovering the tensor decomposition components in all four scenarios, i.e., smaller mean errors, with slightly worse performance in estimating the weight in two out of four scenarios. In terms of the variable selection performance, both our method and the lasso penalized method are able to correctly identify almost all true important variables, while ours tends to include more noisy variables than the lasso method. Next, we compare the computational costs of three methods using scenario **III** for illustration, in which case the tensor has 1 million ($d_1 d_2 d_3 = 10^6$) entries. It takes our Algorithm 1 about 30 seconds to run one replication on a single laptop with 1.3 GHz processor and 4GB memory, which is comparable to the non-sparse tensor decomposition method in Anandkumar et al. (2014c) (30 seconds), and is slower than the lasso penalized method in Allen (2012) (10 seconds). This is partially due to the fact that our TTP method runs $L$ initializations while the lasso penalized method only runs $K$ initialization. Finally, based on the performance of our method across all scenarios, we observe that when dimension is fixed the recovery problem gets harder as $K$ increases. This agrees with our theoretical finding that $\epsilon_R$ increases as $K$ increases.

Table 1: The averaged mean errors, the averaged weight errors, the averaged TPR/FPR variable selection performance in Anandkumar et al. (2014c) (Non-sparse), Allen (2012) (Lasso) and our method (Ours) in all examples of Section 5.2. The standard error is shown in subscript. The minimal error in each scenario is shown in bold.

| Scenarios | Methods | mean error | weight error | TPR | FPR |
|---|---|---|---|---|---|
| I | Non-sparse | $0.295_{0.0218}$ | $0.053_{0.0084}$ | $1_0$ | $1_0$ |
| | Lasso | $0.258_{0.0294}$ | $\mathbf{0.016}_{0.0058}$ | $0.993_{0.0043}$ | $0.009_{0.0034}$ |
| | Ours | $\mathbf{0.171}_{0.0253}$ | $0.021_{0.0053}$ | $0.992_{0.0061}$ | $0.017_{0.0117}$ |
| II | Non-sparse | $0.300_{0.0195}$ | $0.067_{0.0128}$ | $1_0$ | $1_0$ |
| | Lasso | $0.204_{0.0148}$ | $\mathbf{0.008}_{0.0013}$ | $0.998_{0.0008}$ | $0.016_{0.0035}$ |
| | Ours | $\mathbf{0.185}_{0.0224}$ | $0.022_{0.0056}$ | $0.992_{0.0061}$ | $0.086_{0.0215}$ |
| III | Non-sparse | $0.086_{0.0144}$ | $0.015_{0.0101}$ | $1_0$ | $1_0$ |
| | Lasso | $0.055_{0.0029}$ | $\mathbf{0.002}_{0.0004}$ | $1_0$ | $0.003_{0.0022}$ |
| | Ours | $\mathbf{0.036}_{0.0042}$ | $\mathbf{0.002}_{0.0004}$ | $1_0$ | $0.016_{0.0130}$ |
| IV | Non-sparse | $0.196_{0.0416}$ | $0.071_{0.0260}$ | $1_0$ | $1_0$ |
| | Lasso | $0.052_{0.0018}$ | $\mathbf{0.002}_{0.0003}$ | $1_0$ | $0.002_{0.0016}$ |
| | Ours | $\mathbf{0.041}_{0.0064}$ | $\mathbf{0.002}_{0.0003}$ | $1_0$ | $0.067_{0.0311}$ |

# 6 Real Data Analysis

This section investigates the efficacy of our TTP method on Click-Through Rate (CTR) prediction for online advertising. Additional real applications to high dimensional gene clustering will be discussed in Section S.3.2 in the online supplementary materials.

The used advertising dataset comes from a major internet company and forms a third-order tensor (*user-age-group, device-type, advertisement*). Each tensor entry stores the real-valued CTR (ratio of users who click on an ad to the number of total users who see this ad) for the corresponding combination of user-age-group, device-type, and advertisement. The user-age-group consists of 5 categories and the device-type consists of 2 types: PC and mobile. We extracted the most active 100 advertisements that are widely reachable to each user-age-group and each device type. The constructed tensor is of size $5 \times 2 \times 100$. We use the ads click tensor data $\widehat{\mathcal{T}}_1$ on Nov. 1, 2015 for training and use the ads click tensor data $\widehat{\mathcal{T}}_2$ on Nov. 2, 2015 for testing. Since the click event is generally very rare, the constructed tensor is very sparse (about 90% zeros), which indicates that a sparse tensor decomposition method could be suitable for the low rank tensor approximation.

We compare our TTP method with two popular regression methods, linear regression and gradient boosting machine (GBM) (Friedman, 2001). In these two regression methods, the real-valued CTR is treated as the response and (*user-age-group, device-type, advertisement*) are treated as covariates. In our Algorithm 1, given the input tensor $\widehat{\mathcal{T}}_1$, we fix $K = 1$ and the cardinality of (*user-age-group, device-type, advertisement*) as $(5, 2, 10)$ to compute the decomposed components $\widehat{w}, \widehat{\mathbf{a}}, \widehat{\mathbf{b}}, \widehat{\mathbf{c}}$. We evaluate both training error $\|\widehat{\mathcal{T}}_1 - \widehat{w}\widehat{\mathbf{a}} \circ \widehat{\mathbf{b}} \circ \widehat{\mathbf{c}}\|_F$ and testing error $\|\widehat{\mathcal{T}}_2 - \widehat{w}\widehat{\mathbf{a}} \circ \widehat{\mathbf{b}} \circ \widehat{\mathbf{c}}\|_F$. As shown in Table 2, our TTP method greatly improves both regression methods, which indicates a promising potential of our method in exploiting the tensor structure for CTR prediction. The advantage of tensor decomposition method over traditional classification algorithms has also been observed by Rai et al. (2014) and Zhe et al. (2015) for exploiting the binary tensor data.

Table 2: CTR prediction errors via linear regression and generalized boosted regression model (GBM) on flattened tensor data and our sparse tensor decomposition on original tensor data.

| Methods | Training error | Testing error |
|---|---|---|
| Linear regression | 0.189 | 0.534 |
| GBM | 0.190 | 0.533 |
| Ours | **0.141** | **0.511** |

# 7 Discussion

In this paper, we propose a new sparse tensor decomposition method via a truncation step and establish both local and global statistical rates of the TTP estimate. Our method is applied to solve various high-dimensional problems, including high dimensional clustering and mixtures of linear sparse regressions. It is worth noting that our TTP method can also be adapted to solve other high dimensional latent variable models, e.g., mixed membership community detection (Anandkumar et al., 2014a) and dictionary learning (Barak et al., 2014; Peng et al., 2014).

# Supplementary Materials

Will Wei Sun, Junwei Lu, Han Liu, and Guang Cheng

This supplementary contains four parts: (1) Section S.1 explains affiliated steps in our main algorithm; (2) Section S.2 discusses applications of our sparse tensor decomposition to high-dimensional latent variable models; (3) Section S.3 mentions additional simulation study and real data analysis on high-dimensional clustering; (4) Sections S.4 - S.6 are devoted to detailed technical proofs for the theoretical developments.

## S.1 Initialization and Clustering Procedures

This section summarizes the initialization and the clustering procedure in our TTP algorithm. We propose two initialization procedures: one is a sparse SVD initialization and another one is a random initialization. The former has a nice theoretical guarantee, while the latter is simple and fast in practice. We further provide an efficient clustering procedure in order to generate all decomposition components.

### S.1.1 Initialization Step

The first initialization is performed via the sparse SVD initialization, see Algorithm 3. Given a randomly generated Gaussian vector $\boldsymbol{\theta}$, the algorithm first computes its truncated vector $\check{\boldsymbol{\theta}}$, then computes the truncated vectors of the leading left and right singular vectors of the matrix $\widehat{\mathcal{T}} \times_3 \check{\boldsymbol{\theta}}$, and finally outputs the desirable initializations $\widehat{\mathbf{a}}_\tau^{(0)}$ and $\widehat{\mathbf{b}}_\tau^{(0)}$. Given $\widehat{\mathbf{a}}_\tau^{(0)}$ and $\widehat{\mathbf{b}}_\tau^{(0)}$, the vector $\widehat{\mathbf{c}}_\tau^{(0)}$ is computed based on the main update step in (2.6). The input cardinality parameter $(s_1, s_2, s_3)$ in Algorithm 3 is same as the one supplied in Algorithm 1.

**Algorithm 3** Initialization via sparse SVD

1: **Input:** tensor $\widehat{\mathcal{T}}$, cardinality parameter $(s_1, s_2, s_3)$.
2: Step 1: Generate a d-dimensional standard Gaussian vector $\boldsymbol{\theta} \sim N(\mathbf{0}, \mathbf{I}_d)$.
3: Step 2: Compute the sparse vector $\check{\boldsymbol{\theta}} = \text{Truncate}(\boldsymbol{\theta}, \max\{s_1, s_2, s_3\})$.
4: Step 3: Calculate $\mathbf{u}_1$ and $\mathbf{v}_1$ as the leading left and right singular vectors of $\widehat{\mathcal{T}} \times_3 \check{\boldsymbol{\theta}}$.
5: Step 4: Compute the sparse vectors $\check{\mathbf{u}}_1 = \text{Truncate}(\mathbf{u}_1, s_1)$ and $\check{\mathbf{v}}_1 = \text{Truncate}(\mathbf{v}_1, s_2)$.
6: **Output:** $\widehat{\mathbf{a}}_\tau^{(0)} = \text{Norm}(\check{\mathbf{u}}_1)$, $\widehat{\mathbf{b}}_\tau^{(0)} = \text{Norm}(\check{\mathbf{v}}_1)$, and $\widehat{\mathbf{c}}_\tau^{(0)}$ via (2.6) with input $\widehat{\mathbf{a}}_\tau^{(0)}, \widehat{\mathbf{b}}_\tau^{(0)}$.

We provide an intuitive explanation of this initialization procedure. Remind that $\widehat{\mathcal{T}} \times_3 \check{\boldsymbol{\theta}} = \mathcal{T} \times_3 \check{\boldsymbol{\theta}} + \mathcal{E} \times_3 \check{\boldsymbol{\theta}}$ is a multilinear combination of the tensor slices. Based on (2.2), we have $\mathcal{T} \times_3 \check{\boldsymbol{\theta}} = \sum_{i \in [K]} w_i (\mathbf{c}_i^\top \check{\boldsymbol{\theta}}) \mathbf{a}_i \mathbf{b}_i^\top \in \mathbb{R}^{d_1 \times d_2}$. Intuitively, we can treat $w_i(\mathbf{c}_i^\top \check{\boldsymbol{\theta}})$ as the singular value, and $\mathbf{a}_i$,$\mathbf{b}_i$ as the left and right singular vectors. Although this is not an exact singular value decomposition since the spaces of $[\mathbf{a}_1, \ldots, \mathbf{a}_K]$ and $[\mathbf{b}_1, \ldots, \mathbf{b}_K]$ are not orthogonal, we show in Lemma S.4.2 that this algorithm eventually generates good initializations if we repeat this procedure many times. Most importantly, this initialization is shown to lead to the global statistical rate of the final tensor decompositions, see Theorem 3.9.

In practice, in order to save computational cost, we introduce an efficient random initialization procedure. Specifically, we can initialize $\widehat{\mathbf{a}}_\tau^{(0)}$ and $\widehat{\mathbf{b}}_\tau^{(0)}$ as follows. We generate two standard Gaussian vectors, then truncate them by only keeping a few largest absolute values and setting other entries as zeros, and finally normalize them to be of unit norm. The vector $\widehat{\mathbf{c}}_\tau^{(0)}$ is again computed based (2.6), see Algorithm 4. This random initialization is shown to be very efficient in practice, although its theoretical analysis is still unclear.

**Algorithm 4** Random Initialization

1: **Input:** tensor $\widehat{\mathcal{T}}$, cardinality parameter $(s_1, s_2, s_3)$.
2: Step 1: Generate $\mathbf{u}_1$ from $N(\mathbf{0}, \mathbf{I}_{d_1})$ and $\mathbf{v}_1$ from $N(\mathbf{0}, \mathbf{I}_{d_2})$.
3: Step 2: Compute the sparse vectors $\check{\mathbf{u}}_1 = \text{Truncate}(\mathbf{u}_1, s_1)$ and $\check{\mathbf{v}}_1 = \text{Truncate}(\mathbf{v}_1, s_2)$.
4: **Output:** $\widehat{\mathbf{a}}_\tau^{(0)} = \text{Norm}(\check{\mathbf{u}}_1)$, $\widehat{\mathbf{b}}_\tau^{(0)} = \text{Norm}(\check{\mathbf{v}}_1)$, and $\widehat{\mathbf{c}}_\tau^{(0)}$ via (2.6) with input $\widehat{\mathbf{a}}_\tau^{(0)}, \widehat{\mathbf{b}}_\tau^{(0)}$.

### S.1.2 Clustering Step

Algorithm 5 introduces the clustering step in identifying all the $K$ decomposition components from the $L$ generated estimators, see line 7 of Algorithm 1. It outputs the $K$ decompositions sequentially. Within each loop, the algorithm finds the tuple $(\widehat{\mathbf{a}}, \widehat{\mathbf{b}}, \widehat{\mathbf{c}})$ such that $|\widehat{\mathcal{T}} \times_1 \widehat{\mathbf{a}} \times_2 \widehat{\mathbf{b}} \times_3 \widehat{\mathbf{c}}|$ is maximized. The intuition of this step is that if $|\widehat{\mathcal{T}} \times_1 \widehat{\mathbf{a}} \times_2 \widehat{\mathbf{b}} \times_3 \widehat{\mathbf{c}}|$ is large for some $(\widehat{\mathbf{a}}, \widehat{\mathbf{b}}, \widehat{\mathbf{c}})$, then these vectors are close to some true decomposition $(\mathbf{a}_j, \mathbf{b}_j, \mathbf{c}_j), j \in [K]$. Next, the algorithm removes all tuples that are close to $(\widehat{\mathbf{a}}, \widehat{\mathbf{b}}, \widehat{\mathbf{c}})$ since these tuples eventually generate the same decomposition vector up to an error tolerance. This procedure is repeated until all $K$ decompositions are generated.

**Algorithm 5** Clustering procedure

1: **Input:** tensor $\widehat{\mathcal{T}}$, set $S = \{(\widehat{\mathbf{a}}_\tau, \widehat{\mathbf{b}}_\tau, \widehat{\mathbf{c}}_\tau), \tau \in [L]\}$.
2: **For** $j = 1$ **to** $K$ **Do**
3: Step 1: Find $(\widehat{\mathbf{a}}, \widehat{\mathbf{b}}, \widehat{\mathbf{c}}) = \arg\max_{(\mathbf{a},\mathbf{b},\mathbf{c}) \in S} |\widehat{\mathcal{T}} \times_1 \mathbf{a} \times_2 \mathbf{b} \times_3 \mathbf{c}|$.
4: Step 2: Perform $N$ iterations of (2.4)-(2.6) with initialization $(\widehat{\mathbf{a}}, \widehat{\mathbf{b}}, \widehat{\mathbf{c}})$ and denote the final update as $(\widehat{\mathbf{a}}_j, \widehat{\mathbf{b}}_j, \widehat{\mathbf{c}}_j)$.
5: Step 3: Remove all tuples in $S$ with $\min\{\|\widehat{\mathbf{a}}_\tau \pm \widehat{\mathbf{a}}\|, \|\widehat{\mathbf{b}}_\tau \pm \widehat{\mathbf{b}}\|, \|\widehat{\mathbf{c}}_\tau \pm \widehat{\mathbf{c}}\|\} \le 0.5$.
6: **End For**
7: **Output:** $\{(\widehat{\mathbf{a}}_j, \widehat{\mathbf{b}}_j, \widehat{\mathbf{c}}_j), j \in [K]\}$.

In Step 3 of Algorithm 5, the $\pm$ sign takes care of the possibility that $(\widehat{\mathbf{a}}, \widehat{\mathbf{b}}, \widehat{\mathbf{c}})$ may have reverse sign of the true component. The choice of thresholding value 0.5 is not critical. In fact, in our experiments, if the distance between $(\widehat{\mathbf{a}}_\tau, \widehat{\mathbf{b}}_\tau, \widehat{\mathbf{c}}_\tau)$ and $(\widehat{\mathbf{a}}, \widehat{\mathbf{b}}, \widehat{\mathbf{c}})$ is not small, smaller than $10^{-4}$, then their distance will generally be very large, greater than 1. Therefore, setting the thresholding as any constant higher than $10^{-4}$ and smaller than 1 generates the same result.

We next demonstrate the mechanism of the clustering procedure via a simulated example. We consider the same simulation setup as in Section 5.1 and fix $d_1 = d_2 = d_3 = 100, d_{01} = d_{02} = d_{03} = 50$ and $K = 5$. For each $k \in [5]$, we compute the distance of $\min\{\|\widehat{\mathbf{a}}_\tau \pm \widehat{\mathbf{a}}\|, \|\widehat{\mathbf{b}}_\tau \pm \widehat{\mathbf{b}}\|, \|\widehat{\mathbf{c}}_\tau \pm \widehat{\mathbf{c}}\|\}$ in

Step 3 of Algorithm 5. Figure S4 shows the results for $k = 1, 3, 5$, where the thresholding value 0.5 is shown in red in each plot. The cases of $k = 2$ and 4 are similar and hence omitted. As shown in the plots, those initializations close to $(\widehat{\mathbf{a}}, \widehat{\mathbf{b}}, \widehat{\mathbf{c}})$ have distance less than $\epsilon_0 = 10^{-4}$, which is the stopping criterion value used in our algorithm; while those initializations far away from $(\widehat{\mathbf{a}}, \widehat{\mathbf{b}}, \widehat{\mathbf{c}})$ have distance around $\sqrt{2}$. This ensures that the choice of thresholding value 0.5 is not sensitive, which empirically supports our above discussions.

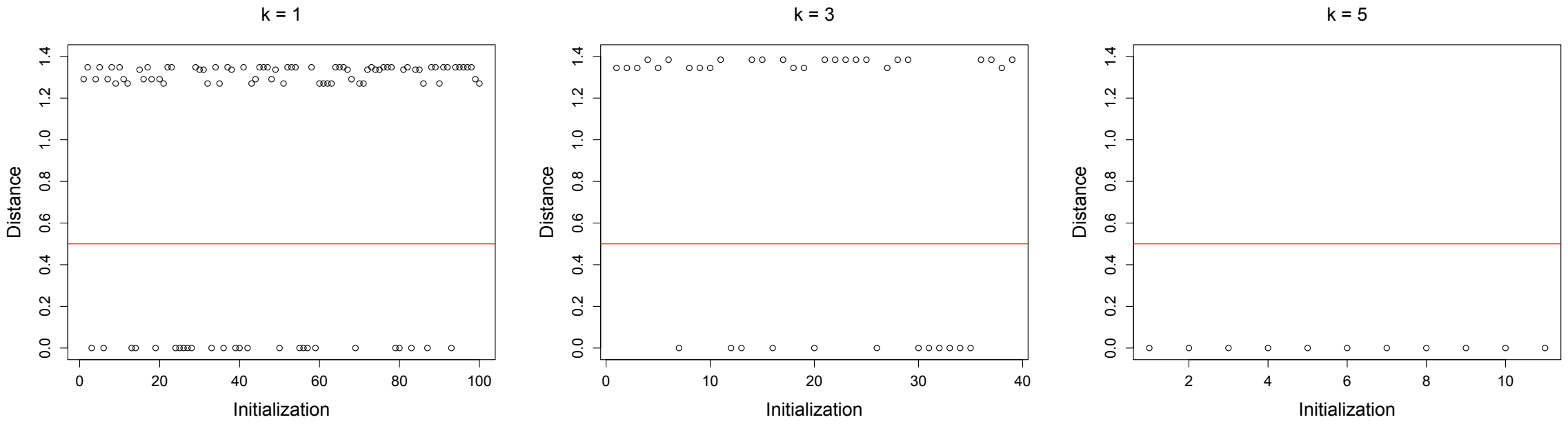

Figure S4: Illustration of the clustering procedure. Each circle represents the distance of each initialization vector to the cluster center. The line in read refers to the thresholding value 0.5 used in our algorithm.

## S.2 Applications of Sparse Tensor Decomposition

This section discusses the applications of our sparse tensor decomposition to high-dimensional latent variable models. Tensor spectral method has been applied to various statistical models with latent variables including Gaussian mixture model (Dasgupta, 1999; Sanjeev and Kannan, 2001; Dasgupta and Schulman, 2007; Vempala and Wang, 2002; Belkin and Sinha, 2010; Kalai et al., 2010; Moitra and Valiant, 2010; Hsu and Kakade, 2013) and mixture linear model (Viele and Tong, 2002; Chaganty and Liang, 2013; Yi et al., 2014). We consider these two models under the high dimensional setting where the dimension is large and the parameters have sparsity structures.

### S.2.1 Sparse Gaussian Mixture Model

The sparse Gaussian mixture model is a generalization of ordinary Gaussian mixture model by assuming that the mixed mean vectors are sparse. In particular, suppose each $d$-dimensional observation $\boldsymbol{x}_i, i = 1, \cdots, n$, is drawn from a Gaussian mixture model with probability density function (pdf) $f(\boldsymbol{x}) = \sum_{k=1}^{K} w_k f_k(\boldsymbol{x}; \boldsymbol{\mu}_k, \boldsymbol{\Sigma}_k)$, where $w_k > 0$ is the mixture weight and $f_k(\boldsymbol{x}; \boldsymbol{\mu}_k, \boldsymbol{\Sigma}_k)$ is the pdf of a multivariate normal distribution,

$$f_k(\boldsymbol{x}; \boldsymbol{\mu}_k, \boldsymbol{\Sigma}_k) = (2\pi)^{-d/2} |\boldsymbol{\Sigma}_k|^{-1/2} \exp\left\{ -\frac{1}{2} (\boldsymbol{x} - \boldsymbol{\mu}_k)^T \boldsymbol{\Sigma}_k^{-1} (\boldsymbol{x} - \boldsymbol{\mu}_k) \right\}.$$

We assume $\|\boldsymbol{\mu}_1\|_0 = \cdots = \|\boldsymbol{\mu}_K\|_0 = d_0$. Notice that the supports of these vectors are not necessarily the same. In addition, to facilitate the high-dimensional clustering as in Pan and Shen (2007), we further assume that a common diagonal covariance matrix is shared among the mixture components. That is, the observations in the $k$-th cluster follow a multivariate Gaussian $N(\boldsymbol{\mu}_k, \sigma^2 \mathbf{I})$.

Our goal is to recover $w_k, \boldsymbol{\mu}_k, \sigma^2$ for $k = 1, \cdots, K$ via our sparse tensor decomposition framework based on the observations $\boldsymbol{x}_i, i = 1, \cdots, n$. The following Lemma relates the model parameters to the second and third moments such that our sparse tensor decomposition can be employed.

**Lemma S.2.1.** (Hsu and Kakade, 2013, Theorem 1) Assume the center matrix $[\boldsymbol{\mu}_1, \ldots, \boldsymbol{\mu}_K]$ has column rank $K$. The variance $\sigma^2$ equals the smallest eigenvalue of the covariance matrix $\mathbb{E}[\boldsymbol{X} \circ \boldsymbol{X}] - \mathbb{E}[\boldsymbol{X}] \circ \mathbb{E}[\boldsymbol{X}]$. Define $\mathcal{M} := \mathbb{E}[\boldsymbol{X} \circ \boldsymbol{X} \circ \boldsymbol{X}] - \sigma^2 \sum_{i=1}^d (\mathbb{E}[\boldsymbol{X}] \circ \mathbf{e}_i \circ \mathbf{e}_i + \mathbf{e}_i \circ \mathbb{E}[\boldsymbol{X}] \circ \mathbf{e}_i + \mathbf{e}_i \circ \mathbf{e}_i \circ \mathbb{E}[\boldsymbol{X}])$, where $\{\mathbf{e}_1, \ldots, \mathbf{e}_d\}$ is the coordinate basis for $\mathbb{R}^d$. Then

$$\mathcal{M} = \sum_{k=1}^K w_k \boldsymbol{\mu}_k \circ \boldsymbol{\mu}_k \circ \boldsymbol{\mu}_k.$$

According to Lemma S.2.1, the model parameters $\sigma^2$ can be recovered based on the empirical covariance matrix, and $w_k, \boldsymbol{\mu}_k$ can be recovered by applying our sparse tensor decomposition procedure to $\widehat{\mathcal{M}}$ estimated from empirical moments. In this scenario, we treat the decomposition components equally, that is, $\mathbf{a}_k = \mathbf{b}_k = \mathbf{c}_k = \boldsymbol{\mu}_k$ for $k \in [K]$.

**Remark S.2.2.** Note that our sparse tensor decomposition procedure assumes unit decomposition vectors, i.e., $\|\boldsymbol{\mu}_k\| = 1$. When the true means $\boldsymbol{\mu}_k$ are not unit, we provide a scaling procedure by first recovering $w_k$ and then recover $\boldsymbol{\mu}_k$ via a similar strategy as in Hsu and Kakade (2013) and Anandkumar et al. (2014b). Denote $\mathbf{M}_2 := \mathbb{E}[\boldsymbol{X} \circ \boldsymbol{X}] - \sigma^2 \mathbf{I}$. Let $\mathbf{U} \in \mathbb{R}^{d \times K}$ be the orthonormal eigenvectors of $\mathbf{M}_2$, and $\mathbf{D} \in \mathbb{R}^{K \times K}$ be the diagonal matrix of positive eigenvalues of $\mathbf{M}_2$. Let $\mathbf{W} := \mathbf{U}\mathbf{D}^{-1/2}$ and $\widetilde{\mathcal{M}} := \mathcal{M}(\mathbf{W}, \mathbf{W}, \mathbf{W})$. We have

$$\widetilde{\mathcal{M}} = \sum_{k=1}^K w_k^{-1/2} \widetilde{\boldsymbol{\mu}}_k \circ \widetilde{\boldsymbol{\mu}}_k \circ \widetilde{\boldsymbol{\mu}}_k,$$

where $\widetilde{\boldsymbol{\mu}}_k := \sqrt{w_k} \mathbf{W}^\top \boldsymbol{\mu}_k$ satisfies $\|\widetilde{\boldsymbol{\mu}}_k\| = 1, k \in [K]$. Note that $\widetilde{\boldsymbol{\mu}}_k \in \mathbb{R}^K$. Therefore, in Step 1 we can apply the robust tensor decomposition method (Anandkumar et al., 2014b) to the non-sparse tensor $\widetilde{\mathcal{M}}$ to obtain the weight $w_k, k \in [K]$. Then the mean vectors $\boldsymbol{\mu}_k$ can be recovered by applying our sparse tensor decomposition to $\mathcal{M}$ and rescaling according to weight $w_k, k \in [K]$ from Step 1. In practice, when $\widehat{\mathbf{M}}_2$ estimated from data is singular, we first apply a thresholding procedure (Bickel and Levina, 2008) on $\widehat{\mathbf{M}}_2$ such that its eigen-decomposition is well defined. When the mean vectors $\boldsymbol{\mu}_k$ are jointly sparse, this thresholding procedure can efficiently shrink the corresponding non-important coordinates in $\widehat{\mathbf{M}}_2$ to be zero.

**Remark S.2.3.** We point out that the success of our sparse tensor decomposition relies on the incoherence condition in Assumption 3.2. When the mean vectors $\boldsymbol{\mu}_k$ does not satisfy this incoherence condition, we show a heuristic procedure to deal with such a situation. The key idea is to shift the original data such that the new cluster means are nearly orthogonal. For example, assume $K = 2$ and the original true mean vectors are $\boldsymbol{\mu}_1 = (1, 1)^\top$ and $\boldsymbol{\mu}_2 = (-1, -1)^\top$. The incoherence condition fails completely since $\boldsymbol{\mu}_1$ and $\boldsymbol{\mu}_2$ are perfectly correlated. By moving the original data toward the direction $(-1,1)^\top$, we obtain two new mean vectors $\widetilde{\boldsymbol{\mu}_1} = (0,2)^\top$ and $\widetilde{\boldsymbol{\mu}}_2 = (-2,0)^\top$, which are orthogonal such that our sparse tensor decomposition method becomes suitable. In practice, the true mean vectors are unknown. We suggest using the standard k-means algorithm to obtain the initial mean vectors and then shift the data such that new cluster centers are nearly orthogonal. In Section S.3.2, we demonstrate that this procedure works very well in practice.

### S.2.2 Mixtures of Linear Sparse Regressions

The mixture of sparse linear model is a generalization of the sparse Gaussian mixture model. It has been employed in music perception, where covariate is the actual tone and response is the tone perceived by a musician (Viele and Tong, 2002). Applying a similar notation system as the Gaussian mixture model, we denote the number of mixture components as $K$. Given the covariates $\boldsymbol{X} \in \mathbb{R}^d$, the mixture of linear sparse regressions is generated as follows.

$(i)$ Draw component label $h \sim \text{Multinomial}(\pi)$, for $\pi = (\pi_1, \ldots, \pi_K)$,

$(ii)$ Draw observation noise $\epsilon$ from a known zero-mean distribution,

$(iii)$ Draw response $Y = \boldsymbol{\beta}_h^\top \boldsymbol{X} + \epsilon$,

where the coefficients $\boldsymbol{\beta}_1, \ldots, \boldsymbol{\beta}_K \in \mathbb{R}^d$ are high-dimensional sparse vectors satisfying the incoherence condition in Assumption 3.2 with $\mathbf{a}_h = \mathbf{b}_h = \mathbf{c}_h = \boldsymbol{\beta}_h$ for $h = 1, \ldots, K$.

In practice, given i.i.d. samples $\mathcal{D} = \{(\boldsymbol{x}_1, y_1), \ldots, (\boldsymbol{x}_n, y_n)\}$ drawn from the mixture of linear regressions, we aim to learn the parameters $\pi, \boldsymbol{\beta}_1, \ldots, \boldsymbol{\beta}_K$. Chaganty and Liang (2013) proposed spectral experts for estimating these parameters via a tensor power method. However, in their approach, the parameters $\boldsymbol{\beta}_1, \ldots, \boldsymbol{\beta}_K$ are not sparse. We combine the spectral experts algorithm in Chaganty and Liang (2013) with the sparse tensor decomposition in this paper and derive a new approach to learn $\pi$ and the sparse coefficients $\boldsymbol{\beta}_1, \ldots, \boldsymbol{\beta}_K$. Denote $\langle \cdot, \cdot \rangle$ as a generalized dot product between two $p$-th order tensors such that $\langle \mathcal{X}, \mathcal{Y} \rangle := \sum_{i_1,\ldots,i_p} \mathcal{X}_{i_1,\ldots,i_p} \mathcal{Y}_{i_1,\ldots,i_p}$.

**Lemma S.2.4.** (Chaganty and Liang, 2013) Define the generated vector and tensor as, respectively, $\mathbf{m} := \sum_{h=1}^K \pi_h \boldsymbol{\beta}_h$ and $\mathcal{M} := \sum_{h=1}^K \pi_h \boldsymbol{\beta}_h \circ \boldsymbol{\beta}_h \circ \boldsymbol{\beta}_h$, then we have

$$Y = \langle \mathbf{m}, \boldsymbol{X} \rangle + \eta_1(\boldsymbol{X}), \quad \text{(S.1)}$$

$$Y^3 = \langle \mathcal{M}, \boldsymbol{X} \circ \boldsymbol{X} \circ \boldsymbol{X} \rangle + 3\mathbb{E}[\epsilon^2] \langle \mathbf{m}, \boldsymbol{X} \rangle + \mathbb{E}[\epsilon^3] + \eta_3(\boldsymbol{X}), \quad \text{(S.2)}$$

where the noise terms $\eta_1(\boldsymbol{X})$ and $\eta_3(\boldsymbol{X})$ have zero means for any covariate $\boldsymbol{X} \in \mathbb{R}^d$.

According to Lemma S.2.4, we can perform two low-rank regressions to recover $\mathbf{m}$ and $\mathcal{M}$, and then apply our sparse tensor decomposition procedure on $\widehat{\mathcal{M}}$ to obtain the estimation for the desirable parameters. Specifically, according to ($S$.1), we first estimate the sparse vector $\mathbf{m}$ via lasso regression (Tibshirani, 1996) with a tuning parameter $\lambda_1 > 0$,

$$\widehat{\mathbf{m}} := \arg\min_{\mathbf{m}} \frac{1}{2n} \sum_{(\boldsymbol{x}_i, \mathbf{y}_i) \in \mathcal{D}} \big( \langle \mathbf{m}, \boldsymbol{x}_i \rangle - \mathbf{y}_i \big)^2 + \lambda_1 \|\mathbf{m}\|_1,$$

and then estimate the tensor $\mathcal{M}$ via a low-rank regression (Chaganty and Liang, 2013)

$$\widehat{\mathcal{M}} := \arg\min_{\mathcal{M}} \frac{1}{2n} \sum_{(\boldsymbol{x}_i, \mathbf{y}_i) \in \mathcal{D}} \Big( \langle \mathcal{M}, \boldsymbol{x}_i \circ \boldsymbol{x}_i \circ \boldsymbol{x}_i \rangle + 3\mathbb{E}[\epsilon^2] \langle \widehat{\mathbf{m}}, \boldsymbol{x}_i \rangle + \mathbb{E}[\epsilon^3] - \mathbf{y}_i^3 \Big)^2 + \lambda_3 \|\mathcal{M}\|_*,$$

where $\lambda_3$ is a tuning parameter and $\|\mathcal{M}\|_* = d^{-1} \sum_{l=1}^{d} \|[\mathcal{M}]_{:,:,l}\|_*$ is the average nuclear norm over all $d$ unfoldings. Finally, we perform our sparse tensor decomposition procedure on the estimator $\widehat{\mathcal{M}}$ to recover the parameters $\pi, \boldsymbol{\beta}_1, \ldots, \boldsymbol{\beta}_K$.

## S.3 Additional Experiments

This section examines the effectiveness of the proposed sparse tensor decomposition method in the high-dimensional clustering problem with both simulated examples and real data analysis.

### S.3.1 Simulation Study

To assess the clustering performance, we define the cluster error (Sun et al., 2012) as the estimated distance between an estimated clustering assignment $\widehat{\psi}$ and the true assignment $\psi$ of the sample data $\boldsymbol{x}_1, \ldots, \boldsymbol{x}_n$.

$$\text{cluster error} = \binom{n}{2}^{-1} \Big| \{(i,j) : \mathbb{1}(\widehat{\psi}(\boldsymbol{x}_i) = \widehat{\psi}(\boldsymbol{x}_j)) \neq \mathbb{1}(\psi(\boldsymbol{x}_i) = \psi(\boldsymbol{x}_j)); i < j\} \Big|,$$

where $|A|$ is the cardinality of set A. In this case, the mean error defined in (5.1) reduces to mean error $= K^{-1} \sum_{k=1}^{K} \big\| \widehat{\boldsymbol{\mu}}_k - \boldsymbol{\mu}_k \big\|$. The weight error, TPR, and FPR can be computed analogously.

The simulated data consist of $n = 1000$ observations $\boldsymbol{x}_i \in \mathbb{R}^d; i \in [n]$ of dimension $d = 10$. First, cluster memberships $\mathbf{y}_i$'s are uniformly sampled from $\{1, 2, 3, 4\}$. Then for each cluster $\mathbf{y}_i$, we generate 250 samples from $N(\boldsymbol{\mu}(\mathbf{y}_i), \sigma^2 \mathbf{I})$, where

$$\boldsymbol{\mu}(\mathbf{y}_i) = \boldsymbol{\mu}_1 \mathbb{1}(\mathbf{y}_i = 1) + \boldsymbol{\mu}_2 \mathbb{1}(\mathbf{y}_i = 2) + \boldsymbol{\mu}_3 \mathbb{1}(\mathbf{y}_i = 3) + \boldsymbol{\mu}_4 \mathbb{1}(\mathbf{y}_i = 4),$$

and $\sigma^2 = 0.1$. To examine the performance in various scenarios, the following four sparse models of mean vector $\boldsymbol{\mu}(\mathbf{y}_i)$ are considered. The final mean vectors are normalized to have unit norm. Denote $\mathbf{e}_i \in \mathbb{R}^d$ as the basis vector whose $i$-th coordinate is 1 and the rests are zeros.

- **Model 1:** $\boldsymbol{\mu}_k$'s are orthonormal and each has $d_0 = 1$ nonzero elements: $\boldsymbol{\mu}_k = \mathbf{e}_k$ for $k \in [4]$.
- **Model 2:** $\boldsymbol{\mu}_k$'s are orthonormal and each has $d_0 = 2$ nonzero elements: $\boldsymbol{\mu}_1 = \mathbf{e}_1 + \mathbf{e}_2, \boldsymbol{\mu}_2 = \mathbf{e}_3 + \mathbf{e}_4, \boldsymbol{\mu}_3 = \mathbf{e}_5 + \mathbf{e}_6, \boldsymbol{\mu}_4 = \mathbf{e}_7 + \mathbf{e}_8$.
- **Model 3:** $\boldsymbol{\mu}_k$'s are not orthogonal with small incoherence parameter $\zeta = 0.2$: $\boldsymbol{\mu}_1 = \mathbf{e}_1 + 0.2\mathbf{e}_2, \boldsymbol{\mu}_2 = \mathbf{e}_2 + 0.2\mathbf{e}_3, \boldsymbol{\mu}_3 = \mathbf{e}_3 + 0.2\mathbf{e}_4, \boldsymbol{\mu}_4 = \mathbf{e}_4 + 0.2\mathbf{e}_1$.
- **Model 4:** $\boldsymbol{\mu}_k$'s are not orthogonal with large incoherence parameter $\zeta = 0.5$: $\boldsymbol{\mu}_1 = \mathbf{e}_1 + 0.9\mathbf{e}_2, \boldsymbol{\mu}_2 = \mathbf{e}_2 + 0.9\mathbf{e}_3, \boldsymbol{\mu}_3 = \mathbf{e}_3 + 0.9\mathbf{e}_4, \boldsymbol{\mu}_4 = \mathbf{e}_4 + 0.9\mathbf{e}_1$.

We evaluate the performance our procedure in reconstruction of the four clusters by taking Model 3 for illustration. Figure S5 shows the original four-cluster samples and the reconstructed samples generated from the distribution based on $\widehat{w}_k, \widehat{\boldsymbol{\mu}}_k, k = 1, \ldots, 4$. Clearly, the reconstructed samples mimic the original observations very well.

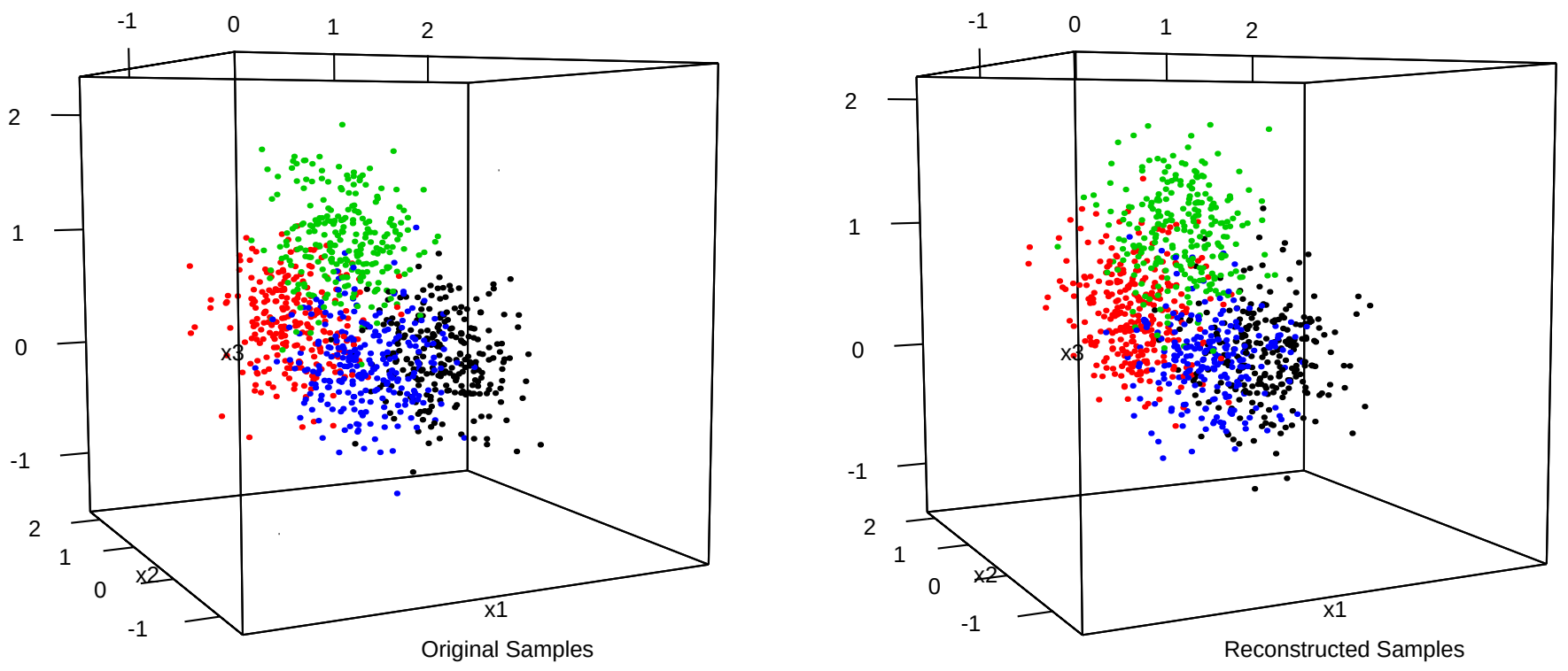

Figure S5: Original samples and reconstructed sample in Model 3 of Section S.3.1.

Finally, we report the cluster errors, the mean estimation errors , the relative estimation errors of weight, as well as the TPR/FPR variable selection performance. We compare two methods: one using empirical tensor $\widehat{\mathcal{M}}$ estimated from samples; another one using true tensor $\mathcal{M}$ based on population parameters. Table S1 summarizes the results. First, our sparse tensor decomposition method successfully identifies the true important variables in all models, which shows the superior variable selection performance. Second, when the true mean vectors are orthonormal, i.e., incoherence parameter $\zeta = 0$ in Models 1-2, using true tensor will fully recover the mean vectors and the weight, while when incoherence parameter $\zeta \neq 0$ in Models 3-4, even decomposing the true tensor can not fully recover them. This observation agrees with our theoretical findings in Theorems 3.6 and 3.9 that the error $\epsilon_R = O(\eta(\mathcal{E}, d_0) + \sqrt{K}/d_0)$ consists of two parts, one is due to sample error $O(\eta(\mathcal{E}, d_0))$ and another one is due to incoherence condition reflected as $O(\sqrt{K}/d_0)$. Third, comparing the mean errors based on $\widehat{\mathcal{M}}$ and $\mathcal{M}$, we observe that the sample error overwhelmingly dominates the model error. In this case, our error rate $\epsilon_R$ significantly improves the rate shown in Anandkumar et al. (2014c), see the discussions after Theorem 3.6.

### S.3.2 Real Data Analysis

We apply the proposed sparse tensor decomposition method to high-dimensional gene clustering for the Leukemia microarray dataset (Golub et al., 1999). The goal is to distinguish two types of human acute leukemias: acute myeloid leukemia(AML) and acute lymphoblastic leukemia(ALL) based on the gene expression data.

This dataset consists of 72 patients in total, 25 patients with AML and 47 patients with ALL. The Gene expression levels were measured by Affymetrix microarrays containing 3571 human genes after preprocessing (Dettling, 2004). Distinguishing ALL from AML is clinically significant for successful treatment because those chemotherapy regimens for ALL patients are different from AML patients, in which case using ALL therapy for AML (and vice versa) cases may result in distinctly reduced cute rates and possible toxicities.

Table S1: The cluster errors, the mean estimation errors, the weight estimation errors, as well as the TPR/FPR variable selection performance of our method in Models 1-4 of Section S.3.1.

| Tensor Type | Model | cluster error | mean error | weight error | TPR | FPR |
|---|---|---|---|---|---|---|
| Estimated $\widehat{\mathcal{M}}$ | Model 1 ($\zeta$=0) | 0.0361 | 0 | 0.0835 | 1 | 0 |
| | Model 2 ($\zeta$=0) | 0.0408 | 0.0318 | 0.0647 | 1 | 0 |
| | Model 3 ($\zeta$=0.2) | 0.0465 | 1.5773 | 0.0924 | 1 | 0 |
| | Model 4 ($\zeta$=0.5) | 0.1189 | 1.0224 | 0.2968 | 1 | 0 |
| True $\mathcal{M}$ | Model 1 ($\zeta$=0) | 0.0361 | 0 | 0 | 1 | 0 |
| | Model 2 ($\zeta$=0) | 0.0408 | 0 | 0 | 1 | 0 |
| | Model 3 ($\zeta$=0.2) | 0.0475 | 0.0797 | 0.0170 | 1 | 0 |
| | Model 4 ($\zeta$=0.5) | 0.1213 | 0.0490 | 0.2470 | 1 | 0 |

Table S2: The selected numbers of informative genes and cluster errors in Leukemia dataset.

| Methods | No. genes | cluster error |
|---|---|---|
| K-means | 3571 | 2/72 |
| Reg. k-means | 211 | 2/72 |
| Ours | **60** | 2/72 |

We compare our method (Ours) with $s = 30$ with the standard k-means algorithm (K-means) and the regularized k-means clustering method (Reg. k-means) in (Sun et al., 2012). As an evaluation criterion, we compare the estimated clustering assignments to the available cancer types of each tumor. The comparison results are summarized in Table S2, where results of k-means and regularized k-means are from Sun et al. (2012). Clearly, our method achieves competitive clustering performance with much less selected important genes compared with the k-means and the regularized k-means clustering algorithms.

## S.4 Proof of Main Results

We provide the proofs of the main results. First we prove the results in Theorem 3.6 for local analysis. Then we establish the results for global analysis in Theorems 3.9. For simplicity, in the following proofs we consider the case $d_1 = d_2 = d_3 = d$ and $d_{01} = d_{02} = d_{03} = d_0$, and hence in our Algorithm 1 we let $s_1 = s_2 = s_3 = s$. The extension of the proofs to a general case with different parameters in three modes is trivial but involves more complicated notations.

### S.4.1 Proof of Theorem 3.6

Our proof consists of two steps. In Step 1, we show a general contraction result in one iteration to quantify the error of $D(\widehat{\mathbf{c}}, \mathbf{c}_j)$ when the input estimators $\widehat{\mathbf{a}}$ and $\widehat{\mathbf{b}}$ satisfying $D(\widehat{\mathbf{a}}, \mathbf{a}_j) \leq \epsilon$ and $D(\widehat{\mathbf{b}}, \mathbf{b}_j) \leq \epsilon$. In Step 2, we carefully calculate the explicit contraction result by applying the assumptions on the perturbation error and initialization. In particular, we show that $D(\widehat{\mathbf{c}}, \mathbf{c}_j) \leq \epsilon_R + q\epsilon_0$, where $\epsilon_R$ is a non-contracting term and $q\epsilon_0$ with $q < 1$ is a contracting term. Then the desirable error bound is obtained by applying this explicit contraction result repeatedly.

**Step 1:** The next lemma accomplished the first step. Define a function $f(\epsilon; K, d_0)$ as

$$f(\epsilon; K, d_0) := \frac{2C_0}{\sqrt{d_0}}\left(1 + C_2\sqrt{\frac{K}{d_0}}\right)^2 \epsilon + C_1\frac{\sqrt{K}}{d_0} + C_3\epsilon^2, \tag{S.1}$$

for some constants $C_0, C_1, C_2, C_3 > 0$. When $K = o(d_0^{3/2})$, the first two terms of $f(\epsilon; K, d_0)$ converge to 0 and the last term is the contracting term.

**Lemma S.4.1.** (General contraction result in one iteration) Consider model in (2.2) satisfying Assumption 3.2, and assume $\|\mathcal{T}\| \leq C_3 w_{\max}$ and $K = o(d_0^{3/2})$. In addition, assume estimators $\widehat{\mathbf{a}}$ in (2.4) and $\widehat{\mathbf{b}}$ in (2.5) of our algorithm satisfy $D(\widehat{\mathbf{a}}, \mathbf{a}_j) \leq \epsilon$ and $D(\widehat{\mathbf{b}}, \mathbf{b}_j) \leq \epsilon$ for some $j \in [K]$. If the perturbation error $\zeta(\mathcal{E}, d_0 + s)$ with $s \geq d_0$ is small enough such that $\zeta(\mathcal{E}, d_0 + s) < w_j(1 - \epsilon^2) - w_{\max} f(\epsilon; K, d_0)$, then the update $\widehat{\mathbf{c}}$ in (2.6) satisfies, with high probability,

$$D(\widehat{\mathbf{c}}, \mathbf{c}_j) \leq \frac{\sqrt{5}w_{\max} f(\epsilon; K, d_0) + \sqrt{5}\zeta(\mathcal{E}, d_0 + s)}{w_j(1 - \epsilon^2) - w_{\max} f(\epsilon; K, d_0) - \zeta(\mathcal{E}, d_0 + s)}. \tag{S.2}$$

If we further assume $D(\widehat{\mathbf{c}}, \mathbf{c}_j) \leq \epsilon$, then the update $\widehat{w} = \widehat{\mathcal{T}} \times_1 \widehat{\mathbf{a}} \times_2 \widehat{\mathbf{b}} \times_3 \widehat{\mathbf{c}}$ satisfies, with high probability, $|\widehat{w} - w_j| \leq 2w_j\epsilon^2 + w_{\max} f(\epsilon; K, d_0) + \zeta(\mathcal{E}, d_0 + s)$.

The detailed proof of Lemma S.4.1 is discussed in Section S.5.1 in the Appendix.

Lemma S.4.1 provides the error bound of one update in a general form. Clearly, when the input error $\epsilon$ increases, the function $f(\epsilon; K, d_0)$ increases and hence the output error bound of $D(\widehat{\mathbf{c}}, \mathbf{c}_j)$ will be larger. Furthermore, if the perturbation error $\zeta(\mathcal{E}, d_0 + s)$ increases, the problem is getting harder and the output error bound will also increase. Moreover, the output error bound in ($S.2$) improves over the input error $\epsilon$ since the only contracting term in $f(\epsilon; K, d_0)$ is in the order of $\epsilon^2$ when $K = o(d_0^{3/2})$.

**Step 2:** In this step, by carefully employing the conditions on the perturbation and initialization, we provide the explicit contract result by simplifying the error bound in ($S.2$). We show that its denominator is lower bounded by $w_{\min}/2$ and hence it is upper bounded by the sum of a contracting term and a constant non-contracting term. Then the desirable error bound follows after $N$ iterations.

Denote $\widetilde{q} := \frac{2C_0}{\sqrt{d_0}}\left(1 + C_2\sqrt{\frac{K}{d_0}}\right)^2 + C_3\epsilon_0$ and $q := \frac{2\sqrt{5}w_{\max}}{w_{\min}}\widetilde{q}$. We have $f(\epsilon_0; K, d_0) = C_1\sqrt{K}/d_0 + \widetilde{q}\epsilon_0$. According to the initialization condition in Assumption 3.5, we have $\widetilde{q} \leq w_{\min}/(4\sqrt{5}w_{\max})$ and hence $q \leq 1/2 < 1$ and $f(\epsilon_0; K, d_0) \leq w_{\min}/(6w_{\max})$. This together with the condition $\zeta(\mathcal{E}, d_0 + s) \leq w_{\min}/6$ and the initialization condition $\epsilon_0 \leq w_{\min}/(6w_{\max})$ implies that the denominator in ($S.2$) has the desirable lower bounded, that is,

$$\begin{aligned} & w_j(1-\epsilon_0^2) - w_{\max} f(\epsilon_0; K, d_0) - \eta(\mathcal{E}, d_0 + s) \\ \geq \ & w_{\min} \left\{ 1 - \frac{w_{\max}}{w_{\min}} \epsilon_0^2 - \frac{w_{\max}}{w_{\min}} f(\epsilon; K, d_0) - \frac{\eta(\mathcal{E}, d_0+s)}{w_{\min}} \right\} \\ \geq \ & \left(1 - \frac{1}{6} - \frac{1}{6} - \frac{1}{6}\right) w_{\min} = \frac{w_{\min}}{2}. \end{aligned} \tag{S.3}$$

This further validates that the assumption $\eta(\mathcal{E}, d_0 + s) < w_j(1-\epsilon_0^2) - w_{\max} f(\epsilon_0; K, d_0)$ in Lemma S.4.1 is fulfilled.

Finally, we bound the whole error term of $D(\widehat{\mathbf{c}}, \mathbf{c}_j)$ by showing that it can be written as a sum of a contracting term and a constant non-contracting term. Specifically, according to ($S.2$) and ($S.3$), in each iteration, we have

$$\begin{aligned} D(\widehat{\mathbf{c}}, \mathbf{c}_j) & \leq \frac{\sqrt{5} w_{\max} f(\epsilon; K, d_0) + \sqrt{5}\eta(\mathcal{E}, d_0+s)}{w_j(1-\epsilon^2) - w_{\max} f(\epsilon; K, d_0) - \eta(\mathcal{E}, d_0+s)} \\ & \leq \frac{2\sqrt{5} C_2 w_{\max}}{w_{\min}} \frac{\sqrt{K}}{d_0} + \frac{2\sqrt{5}}{w_{\min}} \eta(\mathcal{E}, d_0+s) + q\epsilon_0 = \epsilon_R + q\epsilon_0, \end{aligned}$$

where $\epsilon_R$ is a non-contracting constant term and $q\epsilon_0$ is a contracting term. By iteratively applying above inequality, after $N = \Omega(\log\left(\frac{\epsilon_0}{\epsilon_R}\right))$ iterations, we have,

$$\max \left\{ D(\widehat{\mathbf{a}}^{(N)}, \mathbf{a}_j), D(\widehat{\mathbf{b}}^{(N)}, \mathbf{b}_j), D(\widehat{\mathbf{c}}^{(N)}, \mathbf{c}_j) \right\} \leq O(\epsilon_R).$$

The bound of weight $|\widehat{w} - w_j| \leq O(\epsilon_R)$ follows directly. This ends the proof of Theorem 3.6. ■

### S.4.2 Proof of Theorem 3.9

In order to show the global statistical rate of our procedure, we need to first quantify the error bound of the sparse SVD-based initialization in Algorithm 3, and then quantify the accuracy of the clustering process in Algorithm 5. Then the desirable global statistical rate follows by incorporating the local statistical rate result in Theorem 3.6.

The following Lemma establish the error bound of the sparse SVD initialization. The idea is to show that the generated initialization is close to one of the true decomposition components. Define

$$g(L) := \sqrt{2\ln(L)} - \frac{\ln(\ln(L)) + C_0}{2\sqrt{2\ln(L)}} - \sqrt{2\ln(K)},$$

for some positive constant $C_0$, and denote

$$\mu_R = \left(1 + C_2\sqrt{\frac{K}{d_0}}\right)^2, \quad \mu_{\min} := \min\left\{ C_1\sqrt{\frac{K}{d_0}}\left(2 + 2C_2\sqrt{\frac{K}{d_0}} + \frac{C_1}{\sqrt{d_0}}\right), \mu_R \right\},$$

for some positive constants $C_1, C_2$. Let $\mu := (2\mu_R + \widetilde{\mu} - 1)/(1 - \widetilde{\mu})$ for some $0 < \widetilde{\mu} < 1$.

**Lemma S.4.2.** (Sparse SVD initialization) Consider model in (2.2) satisfying Assumption 3.2, and assume $K = O(d_0)$, suppose we run $L$ initialization procedures in Algorithm 3 with $L$ satisfying

$$g(L) \geq \frac{4w_{\max}(1+\mu)\sqrt{\log K}}{w_{\min} - \zeta w_{\max}(1+\mu)},$$

with $\mu < w_{\min}/(\zeta w_{\max}) - 1$, then at least one of the paris $(\widehat{\mathbf{a}}_\tau^{(0)}, \widehat{\mathbf{b}}_\tau^{(0)})$, $\tau \in [L]$, say $j^*$, satisfies

$$\max\left\{D(\widehat{\mathbf{a}}_{j^*}^{(0)}, \mathbf{a}_1), D(\widehat{\mathbf{b}}_{j^*}^{(0)}, \mathbf{b}_1)\right\} \leq \frac{4w_{\max}\mu_{\min}(1+\zeta)\sqrt{\log K} + \alpha_0\sqrt{s}\eta(\mathcal{E}, d_0+s)}{w_{\min}\widetilde{\mu}g(L) - \alpha_0\sqrt{s}\eta(\mathcal{E}, d_0+s)}, \quad \text{(S.4)}$$

with high probability.

The proof of Lemma S.4.2 is discussed in Section S.5.3 in the Appendix. Based on Lemma S.4.2, next we carefully quantify the condition on $L$ such that the error in ($S$.4) satisfies the required initialization condition.

For sufficiently large $d_0$ and when $K = O(d_0)$, it is easy to see that $\gamma$ defined in Assumption 3.5 is lower bounded by $\gamma \geq \min\left\{w_{\min}/12w_{\max}, w_{\min}/(8\sqrt{5}C_3 w_{\max})\right\}$. Setting the upper bound in ($S$.4) as $\min\left\{w_{\min}/12w_{\max}, w_{\min}/(8\sqrt{5}C_3 w_{\max})\right\}$ implies that, it is sufficient to have

$$g(L) = \Omega\left(\frac{1}{\gamma^2}\sqrt{\log K} + \frac{1}{\gamma w_{\min}}\sqrt{s}\eta(\mathcal{E}, d_0+s)\right).$$

According to the condition on perturbation error $\eta(\mathcal{E}, d_0+s) \leq (w_{\min}/C_5)\sqrt{s^{-1}\log K}$, we have $\sqrt{s}\eta(\mathcal{E}, d_0+s)/(\gamma w_{\min}) \leq \sqrt{\log K}/(C_5\gamma)$. When $\gamma$ is small, this term is dominated by $\gamma^{-2}\sqrt{\log K}$. Therefore, it is sufficient to require the number of initialization $L$ satisfies

$$g(L) = \Omega(\gamma^{-2}\sqrt{\log K}), \text{ i.e. }, L = K^{\Omega(1/\gamma^4)}.$$

Next, the justification of the clustering procedure can be adapt from Lemma 17 in Anandkumar et al. (2014c). That is, the clustering process in Algorithm 5 outputs $K$ cluster centers that are $O(\epsilon_R)$ close to the true components of the tensor. Finally, the desirable global statistical rate follows by incorporating Lemma S.4.2 and the local statistical rate result in Theorem 3.6. This ends the proof of Theorem 3.9. ■

## S.5 Proofs of Lemmas S.4.1-S.4.2 and Corollary 3.8

In this section, we present the detailed proofs of Lemma S.4.1 of the general contraction result in one iteration and Lemma S.4.2 of the sparse SVD initialization. In addition, we provide a theoretical justification of the lasso penalized sparse tensor decomposition in Corollary 3.8.

Before that we introduce an important definition to restrict the operation of a tensor on its partial entries. For an index set $F = F_1 \circ F_2 \circ F_3$ with $F_i \subseteq [d]$, we denote $\mathcal{T}_F$ the restriction of the tensor $\mathcal{T}$ on the three modes indexed by $F_1$, $F_2$ and $F_3$, respectively. That is,

$$[\mathcal{T}_F]_{i,j,k} = \begin{cases} [\mathcal{T}]_{i,j,k} & \text{if } i \in F_1, j \in F_2, \text{and } k \in F_3. \\ 0, & \text{otherwise.} \end{cases}$$

### S.5.1 Proof of Lemma S.4.1

We prove it in three stages. In Stage 1, we bound $D(\widetilde{\mathbf{c}}, w_j \mathbf{c}_j)$, where $\widetilde{\mathbf{c}} = \widehat{\mathcal{T}} \times_1 \widehat{\mathbf{a}} \times_2 \widehat{\mathbf{b}}$ denotes the unnormalized and dense update in (2.6); in Stage 2, we bound $D(\widehat{\mathbf{c}}, \mathbf{c}_j)$ for the normalized and sparse update $\widehat{\mathbf{c}}$; in Stage 3, we bound the estimation error of weight update $|\widehat{w} - w_j|$.

**Stage 1:** Denote $F_1 := \text{supp}(\mathbf{a}_j) \cup \text{supp}(\widehat{\mathbf{a}})$, $F_2 := \text{supp}(\mathbf{b}_j) \cup \text{supp}(\widehat{\mathbf{b}})$, and $F_3 := \text{supp}(\mathbf{c}_j) \cup \text{supp}(\check{\mathbf{c}})$, where $\check{\mathbf{c}} = \text{Truncate}(\widetilde{\mathbf{c}}/\|\widetilde{\mathbf{c}}\|, s)$. Let $F := F_1 \circ F_2 \circ F_3$. Consider the following update

$$\bar{\mathbf{c}}' = \frac{\widehat{\mathcal{T}}_F \times_1 \widehat{\mathbf{a}} \times_2 \widehat{\mathbf{b}}}{\|\widehat{\mathcal{T}}_F \times_1 \widehat{\mathbf{a}} \times_2 \widehat{\mathbf{b}}\|}, \tag{S.1}$$

where $\widehat{\mathcal{T}}_F$ denote the restriction of tensor $\widehat{\mathcal{T}}$ on the three modes indexed by $F_1$, $F_2$ and $F_3$. Note that replacing $\bar{\mathbf{c}}$ with $\bar{\mathbf{c}}'$ in (2.6) of our algorithm does not affect the iteration of $\widehat{\mathbf{c}}$ due to the sparsity restriction of $\widehat{\mathcal{T}}_F$ and the scaling-invariant truncation operation. Therefore, in the sequel, we will assume that $\bar{\mathbf{c}}$ is redefined as $\bar{\mathbf{c}}'$, i.e., $\widetilde{\mathbf{c}}$ is redefined as $\widehat{\mathcal{T}}_F \times_1 \widehat{\mathbf{a}} \times_2 \widehat{\mathbf{b}}$, and then bound $D(\widetilde{\mathbf{c}}, w_j \mathbf{c}_j)$.

For two vectors $\mathbf{u}, \mathbf{v} \in \mathbb{R}^d$, the definition of $D(\mathbf{u}, \mathbf{v})$ can be reformulated as $D(\mathbf{u}, \mathbf{v}) = \sup_{\mathbf{z} \perp \mathbf{v}} (\mathbf{z}^\top \mathbf{u})/(\|\mathbf{z}\|\|\mathbf{u}\|)$. When $\|\mathbf{u}\| = \|\mathbf{v}\| = 1$, this reformulation reduces to our original definition. We introduce this reformulation to measure the distance of non-unit vectors later on. Let $\mathbf{z}_a^* \perp \mathbf{a}_j$, $\mathbf{z}_b^* \perp \mathbf{b}_j$ denote the vectors achieving supremum value in the previous formulation of $D(\widehat{\mathbf{a}}, \mathbf{a}_j)$ and $D(\widehat{\mathbf{b}}, \mathbf{b}_j)$. Assume $\|\mathbf{z}_a^*\| = \|\mathbf{z}_b^*\| = 1$. We can decompose $\widehat{\mathbf{a}}$ and $\widehat{\mathbf{b}}$ as

$$\widehat{\mathbf{a}} = \langle \mathbf{a}_j, \widehat{\mathbf{a}} \rangle \mathbf{a}_j + D(\widehat{\mathbf{a}}, \mathbf{a}_j) \mathbf{z}_a^*, \tag{S.2}$$
$$\widehat{\mathbf{b}} = \langle \mathbf{b}_j, \widehat{\mathbf{b}} \rangle \mathbf{b}_j + D(\widehat{\mathbf{b}}, \mathbf{b}_j) \mathbf{z}_b^*. \tag{S.3}$$

Consider any $\mathbf{z}_c \perp \mathbf{c}_j$ with $\|\mathbf{z}_c\| = 1$, let $\widetilde{\mathbf{z}}_c := \text{Truncate}(\mathbf{z}_c, F_3)$, we have

$$\langle \mathbf{z}_c, \widetilde{\mathbf{c}} \rangle = \langle \widetilde{\mathbf{z}}_c, \widetilde{\mathbf{c}} \rangle = \widehat{\mathcal{T}}_F \times_1 \widehat{\mathbf{a}} \times_2 \widehat{\mathbf{b}} \times_3 \widetilde{\mathbf{z}}_c = \mathcal{T}_F \times_1 \widehat{\mathbf{a}} \times_2 \widehat{\mathbf{b}} \times_3 \widetilde{\mathbf{z}}_c + \mathcal{E}_F \times_1 \widehat{\mathbf{a}} \times_2 \widehat{\mathbf{b}} \times_3 \widetilde{\mathbf{z}}_c, \tag{S.4}$$

where the first equality is due to $\text{supp}(\widetilde{\mathbf{c}}) \subseteq F_3$.

According to (*S.2*) and (*S.3*), we can decompose $\mathcal{T}_F \times_1 \widehat{\mathbf{a}} \times_2 \widehat{\mathbf{b}} \times_3 \widetilde{\mathbf{z}}_c$ as follows.

$$\begin{aligned}
\mathcal{T}_F \times_1 \widehat{\mathbf{a}} \times_2 \widehat{\mathbf{b}} \times_3 \widetilde{\mathbf{z}}_c &= \langle \mathbf{a}_j, \widehat{\mathbf{a}} \rangle \langle \mathbf{b}_j, \widehat{\mathbf{b}} \rangle \mathcal{T}_F \times_1 \mathbf{a}_j \times_2 \mathbf{b}_j \times_3 \widetilde{\mathbf{z}}_c + \langle \mathbf{a}_j, \widehat{\mathbf{a}} \rangle D(\mathbf{b}_j, \widehat{\mathbf{b}}) \mathcal{T}_F \times_1 \mathbf{a}_j \times_2 \mathbf{z}_b^* \times_3 \widetilde{\mathbf{z}}_c \\
&\quad + D(\mathbf{a}_j, \widehat{\mathbf{a}}) \langle \mathbf{b}_j, \widehat{\mathbf{b}} \rangle \mathcal{T}_F \times_1 \mathbf{z}_a^* \times_2 \mathbf{b}_j \times_3 \widetilde{\mathbf{z}}_c + D(\mathbf{a}_j, \widehat{\mathbf{a}}) D(\mathbf{b}_j, \widehat{\mathbf{b}}) \mathcal{T}_F \times_1 \mathbf{z}_a^* \times_2 \mathbf{z}_b^* \times_3 \widetilde{\mathbf{z}}_c \\
&= I_1 + I_2 + I_3 + I_4.
\end{aligned}$$

Next we bound each term individually. For the matrices $\mathbf{A}, \mathbf{B}, \mathbf{C}$ defined in (2.2), we have $\mathbf{A}^\top \mathbf{A} = \mathbf{I} + \mathbf{J}_\mathbf{A}$, $\mathbf{B}^\top \mathbf{B} = \mathbf{I} + \mathbf{J}_\mathbf{B}$, and $\mathbf{C}^\top \mathbf{C} = \mathbf{I} + \mathbf{J}_\mathbf{C}$. Assumption 3.2 implies that $\max\{\|\mathbf{J}_\mathbf{A}\|_\infty, \|\mathbf{J}_\mathbf{B}\|_\infty, \|\mathbf{J}_\mathbf{C}\|_\infty\} \le \zeta$. Define $\bar{\mathbf{c}} := \mathbf{C}\text{Diag}(w)$ as the unnormalized matrix, and $\mathbf{J}_\mathbf{A} * \mathbf{J}_\mathbf{B}$ as the Hadamard product (entry-wise multiplication) of $\mathbf{J}_\mathbf{A}$ and $\mathbf{J}_\mathbf{B}$. We have,

$$I_1 \le |\mathcal{T}_F \times_1 \mathbf{a}_j \times_2 \mathbf{b}_j \times_3 \widetilde{\mathbf{z}}_c| = |\mathcal{T} \times_1 \mathbf{a}_j \times_2 \mathbf{b}_j \times_3 \widetilde{\mathbf{z}}_c)| = |\widetilde{\mathbf{z}}_c^\top \bar{\mathbf{c}}_{\setminus j} (\mathbf{J}_\mathbf{A} * \mathbf{J}_\mathbf{B})_j^{\setminus j}| \le C_1 w_{\max} \frac{\sqrt{K}}{d_0},$$

where the first equality is due to Lemma S.6.1 by noting that $\text{supp}(\mathbf{a}_j) \subseteq F_1$, $\text{supp}(\mathbf{b}_j) \subseteq F_2$, and $\text{supp}(\widetilde{\mathbf{z}}_c) \subseteq F_3$, the second equality is due to the fact that for any $\mathbf{u}, \mathbf{v} \in \mathbb{R}^d$, $\mathcal{T} \times_1 \mathbf{u} \times_2 \mathbf{v} =$ $\sum_{i\in[K]} w_i\langle \mathbf{a}_i, \mathbf{u}\rangle\langle \mathbf{b}_i, \mathbf{v}\rangle \mathbf{c}_i = \bar{\mathbf{c}}(\mathbf{A}^\top u * \mathbf{B}^\top \mathbf{v})$ and $\mathbf{z}_c \perp \bar{\mathbf{c}}_j$, and the last inequality is due to Assumption 3.2 and $\|\widetilde{\mathbf{z}}_c\| \leq 1$.

Let $\widetilde{\mathbf{z}}_b^* := \text{Truncate}(\mathbf{z}_b^*, \text{supp}(\mathbf{b}_j))$, we have $\widetilde{\mathbf{z}}_b^* \perp \mathbf{b}_j$ since $\mathbf{z}_b^* \perp \mathbf{b}_j$. Therefore, we have $\mathcal{T}_F \times_1 \mathbf{a}_j \times_2 \mathbf{z}_b^* \times_3 \widetilde{\mathbf{z}}_c = \mathcal{T}_F \times_1 \mathbf{a}_j \times_2 \widetilde{\mathbf{z}}_b^* \times_3 \widetilde{\mathbf{z}}_c$ due to the fact that $\text{supp}(\widetilde{\mathbf{z}}_b^*) \subseteq \text{supp}(\mathbf{b}_j) \subseteq F_2$. Moreover, Lemma S.6.1 implies that $\mathcal{T}_F \times_1 \mathbf{a}_j \times_2 \widetilde{\mathbf{z}}_b^* \times_3 \widetilde{\mathbf{z}}_c = \mathcal{T} \times_1 \mathbf{a}_j \times_2 \widetilde{\mathbf{z}}_b^* \times_3 \widetilde{\mathbf{z}}_c$. Therefore, we have

$$
\begin{aligned}
I_2 &\leq \epsilon|\mathcal{T}_F \times_1 \mathbf{a}_j \times_2 \mathbf{z}_b^* \times_3 \widetilde{\mathbf{z}}_c| = \epsilon|\mathcal{T} \times_1 \mathbf{a}_j \times_2 \widetilde{\mathbf{z}}_b^* \times_3 \widetilde{\mathbf{z}}_c| = \epsilon\left|\widetilde{\mathbf{z}}_c^\top \bar{\mathbf{c}}_{\setminus j}[(\mathbf{J}_\mathbf{A})_j^{\setminus j} * (\mathbf{B}_{\setminus j})^\top \widetilde{\mathbf{z}}_b^*]\right| \\
&\leq \epsilon\|\bar{\mathbf{c}}_{\setminus j}\|\left\|(\mathbf{J}_\mathbf{A})_j^{\setminus j}\right\|_\infty \|\mathbf{B}_{\setminus j}\|\|\widetilde{\mathbf{z}}_b^*\| \leq \frac{C_0 w_{\max}}{\sqrt{d_0}}\left(1 + C_2\sqrt{\frac{K}{d_0}}\right)^2 \epsilon,
\end{aligned}
$$

where the second inequality is because $\|\widetilde{\mathbf{z}}_c\| \leq 1$ and $\|\mathbf{x} * \mathbf{y}\| \leq \|\mathbf{x}\|_\infty \cdot \|\mathbf{y}\|$ for any two vectors $\mathbf{x}, \mathbf{y}$, and the last inequality is due to $\|\widetilde{\mathbf{z}}_b^*\| \leq 1$ and Assumption 3.2.

Similarly, let $\widetilde{\mathbf{z}}_a^* := \text{Truncate}(\mathbf{z}_a^*, \text{supp}(\mathbf{a}_j))$, we can bound $I_3$ via the same strategy as in $I_2$. Moreover, the bound for $I_4$ is desirable by considering the fact $\|\widetilde{\mathbf{z}}_a^*\| \leq 1$, $\|\widetilde{\mathbf{z}}_b^*\| \leq 1$, $\|\widetilde{\mathbf{z}}_c\| \leq 1$, and the assumption on $\|\mathcal{T}\|$. In particular, we can show that

$$
\begin{aligned}
&I_3 \leq \epsilon|\mathcal{T}_F \times_1 \widetilde{\mathbf{z}}_a^* \times_2 \mathbf{b}_j \times_3 \widetilde{\mathbf{z}}_c| = \epsilon|\mathcal{T} \times_1 \widetilde{\mathbf{z}}_a^* \times_2 \mathbf{b}_j \times_3 \widetilde{\mathbf{z}}_c| \leq \frac{C_0 w_{\max}}{\sqrt{d_0}}\left(1 + C_2\sqrt{\frac{K}{d_0}}\right)^2 \epsilon, \\
&I_4 \leq \epsilon^2|\mathcal{T}_F \times_1 \widetilde{\mathbf{z}}_a^* \times_2 \widetilde{\mathbf{z}}_b^* \times_3 \widetilde{\mathbf{z}}_c| = \epsilon^2|\mathcal{T} \times_1 \widetilde{\mathbf{z}}_a^* \times_2 \widetilde{\mathbf{z}}_b^* \times_3 \widetilde{\mathbf{z}}_c| \leq \epsilon^2\|\mathcal{T}\| \leq C_3 w_{\max}\epsilon^2.
\end{aligned}
$$

After we bound all four terms in $\mathcal{T}_F \times_1 \widehat{\mathbf{a}} \times_2 \widehat{\mathbf{b}} \times_3 \widetilde{\mathbf{z}}_c$, our next step is to bound the error term $\mathcal{E}_F \times_1 \widehat{\mathbf{a}} \times_2 \widehat{\mathbf{b}} \times_3 \widetilde{\mathbf{z}}_c$. Then the whole error rate of $\langle \mathbf{z}_c, \widetilde{\mathbf{c}}\rangle$ can be derived based on (S.4). Since $\|\widehat{\mathbf{a}}\|_0 \leq s$, $\|\widehat{\mathbf{b}}\|_0 \leq s$ and $\|\widetilde{\mathbf{z}}_c\|_0 \leq |F_3| \leq d_0 + s$, we have

$$
|\mathcal{E}_F \times_1 \widehat{\mathbf{a}} \times_2 \widehat{\mathbf{b}} \times_3 \widetilde{\mathbf{z}}_c| \quad \leq \quad \left|\mathcal{E}_F \times_1 \widehat{\mathbf{a}} \times_2 \widehat{\mathbf{b}} \times_3 \frac{\widetilde{\mathbf{z}}_c}{\|\widetilde{\mathbf{z}}_c\|}\right| = \left|\mathcal{E} \times_1 \widehat{\mathbf{a}} \times_2 \widehat{\mathbf{b}} \times_3 \frac{\widetilde{\mathbf{z}}_c}{\|\widetilde{\mathbf{z}}_c\|}\right| \leq \eta(\mathcal{E}, d_0 + s),
$$

where the first inequality is due to $\|\widetilde{\mathbf{z}}_c\| \leq 1$, and the first equality is due to Lemma S.6.1.

Combining all above bounds, we have, $\langle \mathbf{z}_c, \widetilde{\mathbf{c}}\rangle \leq w_{\max} f(\epsilon; K, d_0) + \eta(\mathcal{E}, d_0 + s)$. In order to compute the upper bound of the term $D(\widetilde{\mathbf{c}}, w_j\mathbf{c}_j)$, by definition, the rest part is to quantify the lower bound of $\|\widetilde{\mathbf{c}}\|$. By definition, $\widetilde{\mathbf{c}} = \mathcal{T}_F \times_1 \widehat{\mathbf{a}} \times_2 \widehat{\mathbf{b}} + \mathcal{E}_F \times_1 \widehat{\mathbf{a}} \times_2 \widehat{\mathbf{b}}$. According to Lemma S.6.2, for $F = F_1 \circ F_2 \circ F_3$, we have

$$
\begin{aligned}
\mathcal{T}_F \times_1 \widehat{\mathbf{a}} \times_2 \widehat{\mathbf{b}} &= \sum_{i\in[K]} w_i\big\langle \text{Truncate}(\mathbf{a}_i, F_1), \widehat{\mathbf{a}}\big\rangle\big\langle \text{Truncate}(\mathbf{b}_i, F_2), \widehat{\mathbf{b}}\big\rangle \text{Truncate}(\mathbf{c}_i, F_3) \\
&= \sum_{i\in[K]} w_i\langle \mathbf{a}_i, \widehat{\mathbf{a}}\rangle\langle \mathbf{b}_i, \widehat{\mathbf{b}}\rangle \text{Truncate}(\mathbf{c}_i, F_3) \\
&= w_j\langle \mathbf{a}_j, \widehat{\mathbf{a}}\rangle\langle \mathbf{b}_j, \widehat{\mathbf{b}}\rangle \mathbf{c}_j + \sum_{i\neq j} w_i\langle \mathbf{a}_i, \widehat{\mathbf{a}}\rangle\langle \mathbf{b}_i, \widehat{\mathbf{b}}\rangle \text{Truncate}(\mathbf{c}_i, F_3),
\end{aligned}
$$

where the second equality is due to $\text{supp}(\widehat{\mathbf{a}}) \subseteq F_1$, $\text{supp}(\widehat{\mathbf{b}}) \subseteq F_2$, and the last equality is from $\text{supp}(\mathbf{c}_j) \subseteq F_3$. Therefore, we have

$$
\|\widetilde{\mathbf{c}}\| \geq \left\|w_j\langle \mathbf{a}_j, \widehat{\mathbf{a}}\rangle\langle \mathbf{b}_j, \widehat{\mathbf{b}}\rangle \mathbf{c}_j\right\| - \left\|\sum_{i\neq j} w_i\langle \mathbf{a}_i, \widehat{\mathbf{a}}\rangle\langle \mathbf{b}_i, \widehat{\mathbf{b}}\rangle \text{Truncate}(\mathbf{c}_i, F_3)\right\| - \|\mathcal{E}_F \times_1 \widehat{\mathbf{a}} \times_2 \widehat{\mathbf{b}}\|. \tag{S.5}
$$

Since $\langle \mathbf{a}_j, \widehat{\mathbf{a}} \rangle = \sqrt{1 - D(\mathbf{a}_j, \widehat{\mathbf{a}})}$ and $\langle \mathbf{b}_j, \widehat{\mathbf{b}} \rangle = \sqrt{1 - D(\mathbf{b}_j, \widehat{\mathbf{b}})}$, we have the first term is bounded by $\big\| w_j \langle \mathbf{a}_j, \widehat{\mathbf{a}} \rangle \langle \mathbf{b}_j, \widehat{\mathbf{b}} \rangle \mathbf{c}_j \big\| \geq w_j(1-\epsilon^2)$. In addition, according to (S.2) and (S.3), we have

$$\begin{aligned}
& \Big\| \sum_{i \neq j} w_i \langle \mathbf{a}_i, \widehat{\mathbf{a}} \rangle \langle \mathbf{b}_i, \widehat{\mathbf{b}} \rangle \text{Truncate}(\mathbf{c}_i, F_3) \Big\| \\
\leq \;& \Big\| \sum_{i \neq j} w_i \langle \mathbf{a}_i, \mathbf{a}_j \rangle \langle \mathbf{b}_i, \mathbf{b}_j \rangle \mathbf{c}_i \Big\| + \epsilon \Big\| \sum_{i \neq j} w_i \langle \mathbf{a}_i, \mathbf{z}_a^* \rangle \langle \mathbf{b}_i, \mathbf{b}_j \rangle \text{Truncate}(\mathbf{c}_i, F_3) \Big\| \\
& + \epsilon \Big\| \sum_{i \neq j} w_i \langle \mathbf{a}_i, \mathbf{a}_j \rangle \langle \mathbf{b}_i, \mathbf{z}_b^* \rangle \text{Truncate}(\mathbf{c}_i, F_3) \Big\| + \epsilon^2 \Big\| \sum_{i \neq j} w_i \langle \mathbf{a}_i, \mathbf{z}_a^* \rangle \langle \mathbf{b}_i, \mathbf{z}_b^* \rangle \text{Truncate}(\mathbf{c}_i, F_3) \Big\| \\
\leq \;& w_{\max} f(\epsilon; K, d_0),
\end{aligned}$$

where the last inequality is due to the similar strategy when we bound $\langle \mathbf{z}_c, \widetilde{\mathbf{c}} \rangle$. Furthermore, according to Lemma S.6.3, we have $\|\mathcal{E}_F \times_1 \widehat{\mathbf{a}} \times_2 \widehat{\mathbf{b}}\| \leq \|\mathcal{E}_F\| \leq \eta(\mathcal{E}, d_0 + s)$. Plugging these bounds into (S.5) implies that $\|\widetilde{\mathbf{c}}\| \geq w_j(1-\epsilon^2) - w_{\max} f(\epsilon; K, d_0) - \eta(\mathcal{E}, d_0 + s)$. Therefore, when $w_j(1-\epsilon^2) - w_{\max} f(\epsilon; K, d_0) - \eta(\mathcal{E}, d_0+s) > 0$, we have

$$D(\widetilde{\mathbf{c}}, w_j \mathbf{c}_j) \leq \frac{w_{\max} f(\epsilon; K, d_0) + \eta(\mathcal{E}, d_0 + s)}{w_j(1-\epsilon^2) - w_{\max} f(\epsilon; K, d_0) - \eta(\mathcal{E}, d_0 + s)}.$$

**Stage 2:** We next bound $D(\widehat{\mathbf{c}}, \mathbf{c}_j)$ for the normalized and sparse update $\widehat{\mathbf{c}}$. According to the normalization-invariant property of $D(\mathbf{u}, \mathbf{v}) = \sup_{\mathbf{z} \perp \mathbf{v}} (\mathbf{z}^\top \mathbf{u})/(\|\mathbf{z}\|\|\mathbf{u}\|)$, we have $D(\bar{\mathbf{c}}, \mathbf{c}_j) = D(\widetilde{\mathbf{c}}, w_j \mathbf{c}_j)$, where $\bar{\mathbf{c}} = \widetilde{\mathbf{c}}/\|\widetilde{\mathbf{c}}\|$. In addition, denote $F_c$ as the indices of $\bar{\mathbf{c}}$ with the largest $s$ absolute values, by Lemma S.6.4, we have

$$|\text{Truncate}(\bar{\mathbf{c}}, F_c)^\top \mathbf{c}_j| \geq |\bar{\mathbf{c}}^\top \mathbf{c}_j| - \sqrt{\frac{d_0}{s}} \Big(1 + \sqrt{\frac{d_0}{s}}\Big) [1 - (\bar{\mathbf{c}}^\top \mathbf{c}_j)^2], \tag{S.6}$$

where the right-hand side is an increasing function in $|\bar{\mathbf{c}}^\top \mathbf{c}_j|$ for $|\bar{\mathbf{c}}^\top \mathbf{c}_j| \in [0,1]$. According to our algorithm, $\widehat{\mathbf{c}} = \text{Truncate}(\bar{\mathbf{c}}, F_c)/\|\text{Truncate}(\bar{\mathbf{c}}, F_c)\|$. Note that $\|\text{Truncate}(\bar{\mathbf{c}}, F_c)\| \leq 1$ since $\|\bar{\mathbf{c}}\| = 1$ and the set of $F_c$ only keeps part of the entries of $\bar{\mathbf{c}}$. Therefore, we have $D(\widehat{\mathbf{c}}, \mathbf{c}_j) \leq \sqrt{1 - \big(\text{Truncate}(\bar{\mathbf{c}}, F_c)^\top \mathbf{c}_j\big)^2}$. Combining this with (S.6) implies that

$$\begin{aligned}
D(\widehat{\mathbf{c}}, \mathbf{c}_j) &\leq \left\{1 + 2\sqrt{\frac{d_0}{s}}\Big(1 + \sqrt{\frac{d_0}{s}}\Big)\right\}^{1/2} D(\bar{\mathbf{c}}, \mathbf{c}_j) \leq \sqrt{5} D(\bar{\mathbf{c}}, \mathbf{c}_j) && \text{(S.7)} \\
&\leq \frac{\sqrt{5} w_{\max} f(\epsilon; K, d_0) + \sqrt{5}\eta(\mathcal{E}, d_0 + s)}{w_j(1-\epsilon^2) - w_{\max} f(\epsilon; K, d_0) - \eta(\mathcal{E}, d_0 + s)}, && \text{(S.8)}
\end{aligned}$$

where the second inequality is due to $d_0 \leq s$.

**Stage 3:** Finally, we bound $|\widehat{w} - w_j|$. Specifically, decomposing $\widehat{w}$ leads to

$$\widehat{w} = w_j \langle \mathbf{a}_j, \widehat{\mathbf{a}} \rangle \langle \mathbf{b}_j, \widehat{\mathbf{b}} \rangle \langle \mathbf{c}_j, \widehat{\mathbf{c}} \rangle + \sum_{i \neq j} w_i \langle \mathbf{a}_i, \widehat{\mathbf{a}} \rangle \langle \mathbf{b}_i, \widehat{\mathbf{b}} \rangle \langle \text{Truncate}(\mathbf{c}_i, F_3), \widehat{\mathbf{c}} \rangle + \mathcal{E}_F(\widehat{\mathbf{a}}, \widehat{\mathbf{b}}, \widehat{\mathbf{c}}).$$

Using similar strategy as in Stage 1, we can show that

$$\begin{aligned}|w_j - \widehat{w}| &\le \left|w_j\Big(1 - \langle \mathbf{a}_j, \widehat{\mathbf{a}}\rangle\langle \mathbf{b}_j, \widehat{\mathbf{b}}\rangle\langle \mathbf{c}_j, \widehat{\mathbf{c}}\rangle\Big)\right| + |\sum_{i\neq j} w_i\langle \mathbf{a}_i, \widehat{\mathbf{a}}\rangle\langle \mathbf{b}_i, \widehat{\mathbf{b}}\rangle\langle \text{Truncate}(\mathbf{c}_i, F_3), \widehat{\mathbf{c}}\rangle| + |\mathcal{E}_F(\widehat{\mathbf{a}}, \widehat{\mathbf{b}}, \widehat{\mathbf{c}})| \\ &\le 2w_j\epsilon^2 + w_{\max} f(\epsilon; K, d_0) + \eta(\mathcal{E}, d_0 + s).\end{aligned}$$

Combining Stages 1-3 leads to the desirable error rates of decomposition component and the weight estimation. This ends the proof of Lemma S.4.1. ■

### S.5.2 Proof of Corollary 3.8

Since the only difference between Corollary 3.8 and Theorem 3.6 is change the truncation to the soft-thresholding, we only need to show a similar result as (S.8) and all the remaining steps are same as the proof of Theorem 3.6. Let $\mathcal{S}_c$ is the support of the $\mathbf{c}_j$, then $\|\bar{\mathbf{c}}_{\mathcal{S}_c} - \mathbf{c}_{j\mathcal{S}_c}\|_\infty \le D(\bar{\mathbf{c}}, \mathbf{c}_j)/\sqrt{d_0} \le \rho$. Therefore, if $k \in \mathcal{S}_c$, we have $|\mathbf{c}_{jk}| > 2\rho$ and $|\bar{\mathbf{c}}_k| \ge |\mathbf{c}_{jk}| - \|\bar{\mathbf{c}}_{\mathcal{S}_c} - \mathbf{c}_{j\mathcal{S}_c}\|_\infty > \rho$ and $S(\bar{\mathbf{c}}_k, \rho) > 0$. We can further derive that

$$|S(\bar{\mathbf{c}}, \rho)^\top \mathbf{c}_j| \ge |\bar{\mathbf{c}}^\top \mathbf{c}_j| - |S(\bar{\mathbf{c}}, \rho)^\top \mathbf{c}_j - \bar{\mathbf{c}}^\top \mathbf{c}_j| \ge |\bar{\mathbf{c}}^\top \mathbf{c}_j| - \sqrt{d_0}\rho \text{ and}$$

$$D(S(\bar{\mathbf{c}}, \rho), \mathbf{c}_j) \le \sqrt{d_0}\rho + D(\bar{\mathbf{c}}, \mathbf{c}_j) \le \frac{2w_{\max} f(\epsilon; K, d_0) + \eta(\mathcal{E}, 2d_0)}{w_j(1-\epsilon^2) - w_{\max} f(\epsilon; K, d_0) - \eta(\mathcal{E}, 2d_0)}.$$

The remaining part of the proof is same as Theorem 3.6.

### S.5.3 Proof of Lemma S.4.2

The proof idea is outlined as follows. Remind that $\widehat{\mathcal{T}} \times_3 \check{\boldsymbol{\theta}} = \mathcal{T} \times_3 \check{\boldsymbol{\theta}} + \mathcal{E} \times_3 \check{\boldsymbol{\theta}}$ is a multilinear combination of the tensor slices. Based on (2.2), we have $\mathcal{T} \times_3 \check{\boldsymbol{\theta}} = \sum_{i\in[K]} w_i(\mathbf{c}_i^\top \check{\boldsymbol{\theta}})\mathbf{a}_i\mathbf{b}_i^\top \in \mathbb{R}^{d\times d}$ when $d_1 = d_2 = d_3 = d$. Intuitively, we can treat $w_i(\mathbf{c}_i^\top \check{\boldsymbol{\theta}})$ as the singular value, and $\mathbf{a}_i$,$\mathbf{b}_i$ as the left and right singular vectors. Although this is not an exact singular value decomposition since the spaces of $[\mathbf{a}_1, \dots, \mathbf{a}_K]$ and $[\mathbf{b}_1, \dots, \mathbf{b}_K]$ are not orthogonal, we can show that this sparse SVD algorithm eventually generates good initializations if we repeat this procedure many times.

For the vector $\boldsymbol{\theta} \sim N(\mathbf{0}, \mathbf{I}_d)$ in Algorithm 3, let $\mathbf{u}_1$ and $\mathbf{v}_1$ be the leading left and right singular vectors of $\widehat{\mathcal{T}} \times_3 \check{\boldsymbol{\theta}}$. Let $\lambda := \text{Diag}(w)\mathbf{C}^\top \boldsymbol{\theta} \in \mathbb{R}^k$, and $\lambda_1 := \max_i |\lambda_i|$ and $\lambda_2 := \max_{i\neq 1} |\lambda_i|$. Moreover, denote $\lambda^{(\tau)} = \text{Diag}(w)\mathbf{C}^\top \boldsymbol{\theta}_\tau$. Recall that $\zeta$ is defined in Assumption 3.2. Adapted from Lemma 8 in Anandkumar et al. (2014c), we have, with high probability,

$$\lambda_1^{(j^*)} \ge w_{\min} g(L) \text{ and } \lambda_2^{(j^*)} \le 4w_{\max}(1+\eta)\sqrt{\log K}. \tag{S.9}$$

Let the set of support $\widetilde{F} := \{\text{supp}(\mathbf{a}_1) \cup \text{supp}(\check{\mathbf{u}}_1)\} \circ \{\text{supp}(\mathbf{b}_1) \cup \text{supp}(\check{\mathbf{v}}_1)\} \circ \{\text{supp}(\mathbf{c}_1) \cup \text{supp}(\check{\boldsymbol{\theta}})\}$, We can show that the algorithm has the same output if we replace $\widehat{\mathcal{T}} \times_3 \check{\boldsymbol{\theta}}$ with $\widehat{\mathcal{T}}_{\widetilde{F}} \times_3 \check{\boldsymbol{\theta}}$ in Algorithm 3. Decomposing $\widehat{\mathcal{T}}_{\widetilde{F}} \times_3 \check{\boldsymbol{\theta}} = \mathcal{T}_{\widetilde{F}} \times_3 \check{\boldsymbol{\theta}} + \mathcal{E}_{\widetilde{F}} \times_3 \check{\boldsymbol{\theta}}$, and adapting Lemmas 8-9 in Anandkumar et al. (2014c) on $\widetilde{F}$, we have

$$\max\{D(\mathbf{u}_1, \mathbf{a}_1), D(\mathbf{v}_1, \mathbf{b}_1)\} \le \frac{\mu_{\min}\lambda_2 + \|\mathcal{E}_{\widetilde{F}} \times_3 \check{\boldsymbol{\theta}}\|}{\widetilde{\boldsymbol{\mu}}\lambda_1 - \|\mathcal{E}_{\widetilde{F}} \times_3 \check{\boldsymbol{\theta}}\|}.$$

Let $\widetilde{\boldsymbol{\theta}} = \check{\boldsymbol{\theta}}/\|\check{\boldsymbol{\theta}}\|$, then we have $\|\mathcal{E}_{\widetilde{F}} \times_3 \check{\boldsymbol{\theta}}\| \le \|\check{\boldsymbol{\theta}}\|\|\mathcal{E}_{\widetilde{F}} \times_3 \widetilde{\boldsymbol{\theta}}\| \le \|\check{\boldsymbol{\theta}}\|\|\mathcal{E}_{\widetilde{F}}\| \le \|\check{\boldsymbol{\theta}}\|\eta(\mathcal{E}, d_0 + s)$ where the second inequality is due to $\|\widetilde{\boldsymbol{\theta}}\| = 1$. Next we bound $\|\check{\boldsymbol{\theta}}\|$. Note that the vector $\check{\boldsymbol{\theta}}$ consists of $d_0$ i.i.d. standard normals and the rest are all zeros. According to Lemma S.6.5, we have $P[\|\check{\boldsymbol{\theta}}\| \ge \alpha_0\sqrt{s}] \le e^{-(\alpha_0-1)^2 d_0/2}$. Therefore,

$$P\Big[\|\mathcal{E}_{\widetilde{F}} \times_3 \check{\boldsymbol{\theta}}\| \le \alpha_0\sqrt{s}\eta(\mathcal{E}, d_0 + s)\Big] \ge 1 - e^{-(\alpha_0-1)^2 d_0/2}.$$

This together with (S.9) leads to the desirable result, that is, if

$$g(L) \ge \frac{w_{\max}(1+\mu)}{w_{\min} - \zeta w_{\max}(1+\mu)} 4\sqrt{\log K},$$

with $\mu < w_{\min}/(\zeta w_{\max}) - 1$ and some $0 < \widetilde{\mu} < 1$, then with high probability,

$$\max\Big\{D(\widehat{\mathbf{a}}_{j^*}^{(0)}, \mathbf{a}_1), D(\widehat{\mathbf{b}}_{j^*}^{(0)}, \mathbf{b}_1)\Big\} \le \frac{4w_{\max}\mu_{\min}(1+\zeta)\sqrt{\log K} + \alpha_0\sqrt{s}\eta(\mathcal{E}, d_0+s)}{w_{\min}\widetilde{\mu}g(L) - \alpha_0\sqrt{s}\eta(\mathcal{E}, d_0+s)}.$$

This ends the proof of Lemma S.4.2. ■

## S.6 Auxiliary Lemmas

The following auxillary lemmas S.6.1-S.6.4 are useful to show the general contraction result in one iteration, and Lemma S.6.5 on the tail bound for chi-squared variable is useful when we show the error bound of the sparse SVD-based initialization.

**Lemma S.6.1.** For any tensor $\mathcal{T} \in \mathbb{R}^{d\times d\times d}$ and an index set $F = F_1 \circ F_2 \circ F_3$ with $F_i \subseteq \{1, \dots, d\}$, for any vectors $\mathbf{x}, \mathbf{y}, \mathbf{z} \in \mathbb{R}^d$, if $\text{supp}(\mathbf{x}) \subseteq F_1$, $\text{supp}(\mathbf{y}) \subseteq F_2$, and $\text{supp}(\mathbf{z}) \subseteq F_3$, we have

$$\mathcal{T}_F \times_1 \mathbf{x} \times_2 \mathbf{y} \times_3 \mathbf{z} = \mathcal{T} \times_1 \mathbf{x} \times_2 \mathbf{y} \times_3 \mathbf{z}.$$

**Proof of Lemma S.6.1:** By definition, we get $\mathcal{T}_F \times_1 \mathbf{x} \times_2 \mathbf{y} \times_3 \mathbf{z} = \sum_{i,j,k\in[d]} \mathbf{x}_i\mathbf{y}_j\mathbf{z}_k[\mathcal{T}_F]_{i,j,k}$. Since $[\mathcal{T}_F]_{i,j,k} \neq 0$ only when $i \in F_1, j \in F_1$ and $k \in F_3$, we have $\mathcal{T}_F \times_1 \mathbf{x} \times_2 \mathbf{y} \times_3 \mathbf{z} = \sum_{i\in F_1, j\in F_2, k\in F_3} \mathbf{x}_i\mathbf{y}_j\mathbf{z}_k[\mathcal{T}]_{i,j,k}$. Due to the assumption that $\text{supp}(\mathbf{x}) \subseteq F_1$, $\text{supp}(\mathbf{y}) \subseteq F_2$, and $\text{supp}(\mathbf{z}) \subseteq F_3$, we get the desirable result,

$$\mathcal{T}_F \times_1 \mathbf{x} \times_2 \mathbf{y} \times_3 \mathbf{z} = \sum_{i\in F_1, j\in F_2, k\in F_3} \mathbf{x}_i\mathbf{y}_j\mathbf{z}_k[\mathcal{T}]_{i,j,k} = \sum_{i,j,k\in[d]} \mathbf{x}_i\mathbf{y}_j\mathbf{z}_k[\mathcal{T}]_{i,j,k} = \mathcal{T} \times_1 \mathbf{x} \times_2 \mathbf{y} \times_3 \mathbf{z}.$$

**Lemma S.6.2.** For any tensor $\mathcal{T} \in \mathbb{R}^{d\times d\times d}$ and an index set $F = F_1 \circ F_2 \circ F_3$ with $F_i \subseteq \{1, \dots, d\}$, if $\mathcal{T} = \sum_{i\in[K]} w_i\mathbf{a}_i \circ \mathbf{b}_i \circ \mathbf{c}_i$, we have

$$\mathcal{T}_F = \sum_{i\in[K]} w_i \text{Truncate}(\mathbf{a}_i, F_1) \circ \text{Truncate}(\mathbf{b}_i, F_2) \circ \text{Truncate}(\mathbf{c}_i, F_3).$$

This is a direct application of the sparsity of $\mathcal{T}_F$ and hence the proof is omitted.

**Lemma S.6.3.** For any tensor $\mathcal{T} \in \mathbb{R}^{d\times d\times d}$ and any vectors $\mathbf{x}, \boldsymbol{y} \in \mathbb{R}^d$ with $\|\mathbf{x}\| = \|\mathbf{y}\| = 1$, we have $\|\mathcal{T} \times_1 \mathbf{x} \times_2 \mathbf{y}\| \le \|\mathcal{T}\|$.

**Proof of Lemma S.6.3:** By definition of tensor norm and the property $\|\mathbf{x}\| = \|\mathbf{y}\| = 1$, we have

$$\begin{aligned} \|\mathcal{T}\| &= \sup_{\|\mathbf{u}\|=\|\mathbf{v}\|=\|\mathbf{w}\|=1} |\mathcal{T} \times_1 \mathbf{u} \times_2 \mathbf{v} \times_3 \mathbf{w}| \geq \sup_{\|\mathbf{w}\|=1} |\mathcal{T} \times_1 \mathbf{x} \times_2 \mathbf{y} \times_3 \mathbf{w}| \\ &= \sup_{\|\mathbf{w}\|=1} |\mathbf{w}^\top \mathcal{T} \times_1 \mathbf{x} \times_2 \mathbf{y}| = \|\mathcal{T} \times_1 \mathbf{x} \times_2 \mathbf{y}\|, \end{aligned}$$

where the last equality is due to the fact that for any vector $\mathbf{z} \in \mathbb{R}^d$, $\frac{|\mathbf{w}^\top \mathbf{z}|}{\|\mathbf{w}\|\|\mathbf{z}\|} \leq 1$.

**Lemma S.6.4.** (Yuan and Zhang, 2013, Lemma 12) Consider a sparse vector $\mathbf{x}$ with $\text{supp}(\mathbf{x}) = F_{\mathbf{x}}$ and $|F_{\mathbf{x}}| = d_0$. Let $F_y = \text{supp}(\mathbf{y}, s)$. If $\|\mathbf{x}\| = \|\mathbf{y}\| = 1$, then

$$|\text{Truncate}(\mathbf{y}, F_{\mathbf{y}})^\top \mathbf{x}| \geq |\mathbf{y}^\top \mathbf{x}| - \sqrt{\frac{d_0}{s}} \min\left\{ \sqrt{1 - (\mathbf{y}^\top \mathbf{x})^2}, \left(1 + \sqrt{\frac{d_0}{s}}\right)(1 - (\mathbf{y}^\top \mathbf{x})^2) \right\}.$$

**Lemma S.6.5.** (Laurent and Massart, 2000, Lemma 1) For a Chi-squared variable $Y \sim \chi^2(d)$, we have, for all $x > 0$, $\mathbb{P}[Y - d \geq 2\sqrt{dx} + 2x] \leq e^{-x}$.